\documentclass[runningheads]{llncs}
\usepackage{graphicx}
\usepackage{comment}
\usepackage{amsmath,amssymb} %
\usepackage{color}

\usepackage[width=122mm,left=12mm,paperwidth=146mm,height=193mm,top=12mm,paperheight=217mm]{geometry}
\usepackage{epsfig}
\usepackage{graphicx}
\usepackage{amsmath}
\usepackage{amssymb} 
\usepackage{footnote}
\usepackage[utf8]{inputenc}

\usepackage{graphicx}
\usepackage{color}
\usepackage{xspace}
\usepackage{paralist} 
\usepackage{import}
\usepackage{array,booktabs,calc}
\usepackage{placeins}
\usepackage{cite}
\usepackage{url} 
\renewcommand*{\thefootnote}{\fnsymbol{footnote}}

\newcommand{\bc}{\mathbf{c}}

\newcommand{\bff}{\mathbf{f}} %

\newcommand{\bI}{\mathbf{I}}

\newcommand{\bl}{\mathbf{l}}

\newcommand{\bn}{\mathbf{n}}

\newcommand{\bp}{\mathbf{p}}

\newcommand{\br}{\mathbf{r}}
\newcommand{\bs}{\mathbf{s}}

\newcommand{\bv}{\mathbf{v}}

\newcommand{\bz}{\mathbf{z}}

\newcommand{\nR}{\mathbb{R}}

\newcommand{\cB}{\mathcal{B}}

\newcommand{\cL}{\mathcal{L}}

\newcommand{\cN}{\mathcal{N}}

\newcommand{\cS}{\mathcal{S}}

\newcommand{\cZ}{\mathcal{Z}}

\newcommand{\figref}[1]{Fig.~\ref{#1}}
\newcommand{\secref}[1]{Section~\ref{#1}}

\newcommand{\eqnref}[1]{Eq.~\eqref{#1}}
\newcommand{\tabref}[1]{Table~\ref{#1}}

\makeatletter
\DeclareRobustCommand\onedot{\futurelet\@let@token\@onedot}
\def\@onedot{\ifx\@let@token.\else.\null\fi\xspace}
\def\eg{e.g\onedot} 
\def\ie{i.e\onedot} 
 
\def\etc{etc\onedot}

\def\wrt{wrt\onedot}

\def\etal{et~al\onedot}

\makeatother

\newcommand{\boldparagraph}[1]{\vspace{0.2cm}\noindent{\bf #1:} }

\definecolor{darkgreen}{rgb}{0,0.7,0}
\definecolor{orange}{rgb}{1,0.5,0}

\usepackage{rotating}

\begin{document}
\pagestyle{headings}
\mainmatter

\title{Learning Implicit Surface Light Fields}
\author{Michael Oechsle$^{1,2,3}$ \hspace{0.3cm} Michael Niemeyer$^{1,2}$ \hspace{0.3cm} Lars Mescheder$^{1,2,4\star}$\\
Thilo Strauss$^{3}$ \hspace{0.3cm} Andreas Geiger$^{1,2}$\\}
\institute{$^1$Max Planck Institute for Intelligent Systems, Tübingen \qquad $^2$University of Tübingen \\
$^3$ETAS GmbH, Bosch Group, Stuttgart \qquad $^4$Amazon, Tübingen\\
{\tt\small \{firstname.lastname\}@tue.mpg.de}
{\tt\small $^{3}$\{firstname.lastname\}@etas.com}}
\authorrunning{Oechsle et al.}

\maketitle

\footnotetext[1]{This work was done prior to joining Amazon.}
\renewcommand*{\thefootnote}{\arabic{footnote}}
\setcounter{footnote}{0}

\begin{abstract}
Implicit representations of 3D objects have recently
achieved impressive results on learning-based 3D reconstruction tasks.
While existing works use simple texture models to represent object appearance, photo-realistic image synthesis requires reasoning about the complex interplay of light, geometry and surface properties.
In this work, we propose a novel implicit representation for capturing the visual appearance of an object in terms of its surface light field.
In contrast to existing representations, our implicit model represents surface light fields in a continuous fashion and independent of the geometry.
Moreover, we condition the surface light field with respect to the location and color of a small light source.
Compared to traditional surface light field models, this allows us to manipulate the light source and relight the object using environment maps.
We further demonstrate the capabilities of our model to predict the visual appearance of an unseen object from a single real RGB image and corresponding 3D shape information.
As evidenced by our experiments, our model is able to infer rich visual appearance including shadows and specular reflections.
Finally, we show that the proposed representation can be embedded into a variational auto-encoder for generating novel appearances that conform to the specified illumination conditions.

\end{abstract}

\section{Introduction}
\label{Sec:Intro}

\begin{figure}
\includegraphics[width=\linewidth]{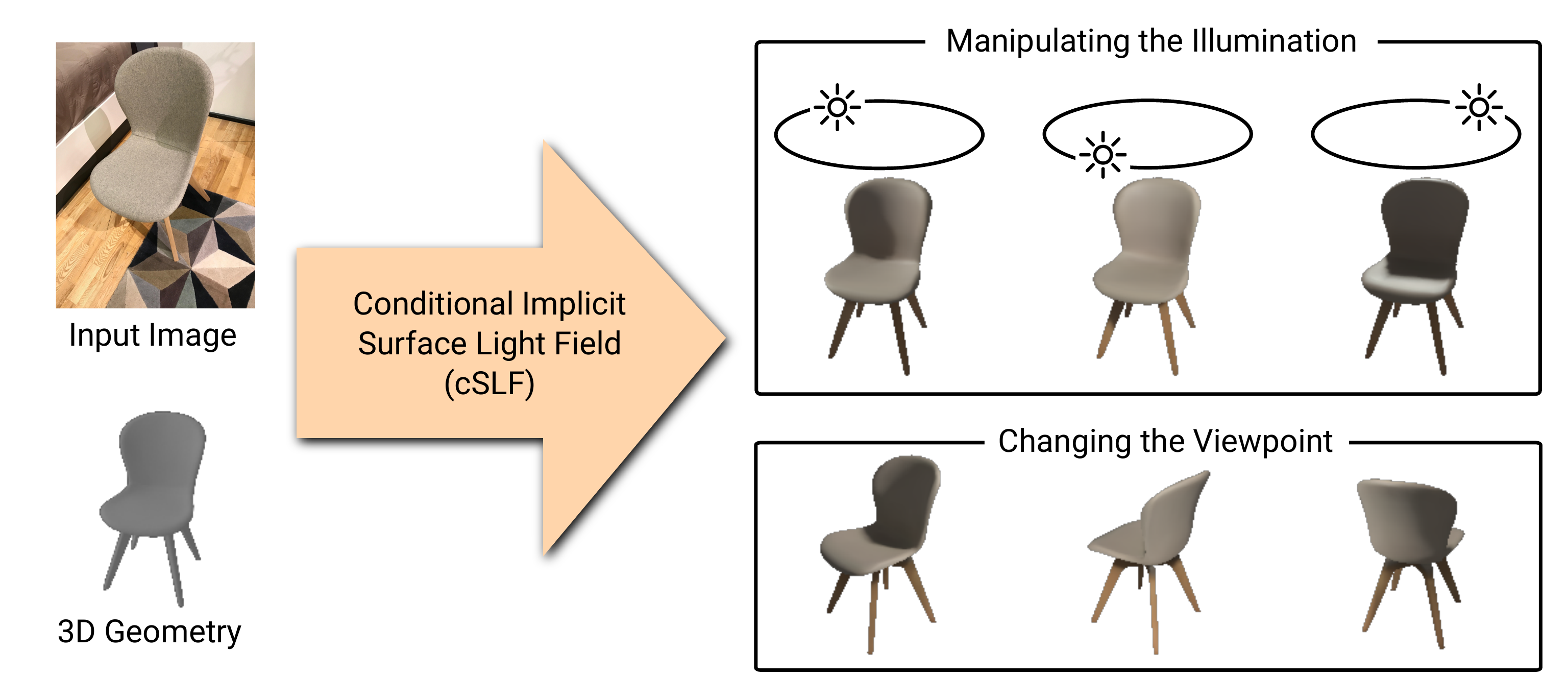}
\caption{\textbf{Illustration.} Given an RGB image and corresponding (predicted or true) 3D geometry as input, our conditional implicit surface light field model allows for manipulating the light configuration and synthesizing novel viewpoints. Note how shadows and specular reflections are faithfully captured by our model.}
\label{fig:teaser}
\end{figure}

Recently, neural implicit representations of 3D objects have emerged as a powerful paradigm for reasoning about the geometry and texture of objects from a single image as input~\cite{Mescheder2019CVPR, Oechsle2019ICCV, Park2019CVPR, Niemeyer2019ARXIV, Sitzmann2019NIPS, Saito2019ARXIV}.
The main advantage of implicit models is their ability to represent 3D structure continuously while handling arbitrary geometric topologies.
Unfortunately, however, existing implicit methods are not able to model the full visual appearance of 3D objects which requires reasoning about the complex interplay of light, geometry and surface properties.
Consequently, the resulting 3D reconstructions appear ``lifeless'', missing view- and light-dependent illumination effects such as shadows or specular reflections.

In the graphics and vision community, synthesizing rich visual appearance of objects using learning-based methods has emerged as a popular research direction with the promise to replace the currently dominating, but slow and work-intensive 3D modeling and physically-based rendering \cite{Jakob2010} paradigm.
Some methods learn neural rendering of voxel features \cite{Sitzmann2019CVPR, Nguyen-Phuoc2018NIPS} or neural texture maps \cite{Thies2019SIG} for synthesizing novel views of an object.
Similarly, Chen \etal \cite{Chen2018SIG} learn the surface light field of a single object.
Even though these methods achieve high realism, they must be retrained for each new object and they do not allow modification of the illumination. In contrast, in this work, we are interested in the following, more challenging scenario: Our goal is to infer \textit{conditional} surface light fields which allow for manipulating illumination conditions and which can be applied to novel, previously unseen objects.

More specifically, we propose an implicit representation for parameterizing conditional surface light fields. As illustrated in \figref{fig:teaser}, our model takes an RGB image and corresponding 3D shape information of an object as input and allows for generating photo-realistic images of the object from arbitrary viewpoints and light configurations.
Towards this goal, we train our model to regress color values given a 3D location on the object's surface, the camera viewpoint and the location and color of a point light source.
We empirically demonstrate that our network represents high-frequency material properties while generalizing across object shapes, light settings and viewpoints.
While trained on synthetic renderings and ground truth geometry, we find that our method generalizes to real images and captures complex physical illumination phenomena including shadows and specular highlights.
Exploiting the well-known rendering equation \cite{Kajiya1986SIGGRAPH}, we demonstrate that our model extends beyond point light sources, enabling implicit rendering using realistic environment maps.
We also extend our representation to the generative setting which allows for synthesizing novel appearances conforming to given illumination conditions and for transferring appearances from one model to another.
Our contributions are summarized as follows:
\begin{itemize}
    \item We propose \textbf{Conditional Implicit Surface Light Fields (cSLF)}, a novel appearance representation of 3D objects which allows for rendering novel views and varying the light color and the light location.
    \item We experimentally verify that our method is able to represent textures, diffuse and specular reflections as well as shadows of a 3D object.
    \item We apply our model for predicting the appearance of a novel object from a single RGB image given ground truth and inferred 3D shape information.
    \item We demonstrate the generative modeling capabilities of our representation by transferring and synthesizing novel physically plausible appearances.
\end{itemize}

\section{Related Work}
\label{Sec:RelWork}

We now discuss the most related works on neural and differentiable rendering, surface light field estimation and implicit shape and appearance modeling.

\boldparagraph{Neural Rendering}
The recent success of Generative Adversarial Networks (GANs) \cite{Goodfellow2014NIPS, Brock2019ICLR, Karras2018ICLR, Karras2019CVPR} has enabled novel image synthesis approaches which employ 2D or 3D convolutional neural networks to generate image content directly from a latent code.
These methods mainly differ in terms of the input representation.
Examples include voxel-based representations \cite{Nguyen-Phuoc2018NIPS, Sitzmann2019CVPR, Rematas2019ARXIV, Nguyen-Phuoc2019ICCV, Nguyen-Phuoc2020ARXIV}, primitive-based representations \cite{Liao2019ARXIV}, depth-based representations \cite{Alhaija2018ACCV, Zhu2018NIPS} and texture-based representations \cite{Thies2019SIG}.
In contrast to CNN based rendering approaches, our method is inherently 3D consistent as we directly predict the object appearance in 3D.
Moreover, our model allows for fine grained viewpoint and lighting control.

\boldparagraph{Differentiable Rendering}
Recently, several methods proposed to backpropagate gradients through the rendering pass \cite{Loper2014ECCV, Chen2019NIPS, Kato2018CVPR, Liu2019ICCV}.
This allows for predicting explicit 3D appearance representations including texture maps \cite{Chen2019NIPS, Kato2018CVPR, Liu2019ICCV} and voxel colors \cite{Henzler2019ICCV}.
Chen \etal \cite{Chen2019NIPS} learn to predict the geometry and texture of a 3D mesh as well as the light position from a single image.
In contrast to these methods, we are interested in inferring a conditional surface light field which also captures more complex physical light-transport phenomena including both shadows and specular highlights. Furthermore, our implicit representation is continuous and independent of the underlying geometric representation.

\boldparagraph{Surface Light Fields}
A surface light field represents the outgoing light at a given location on a 3D surface as a function of the viewing direction \cite{Gortler1996SIG ,Wood2000SIG}.
Maximov \etal \cite{Maximov2019ICCV} propose a method for representing material-light interaction with a neural network which maps normals and the viewing direction to color values.
Chen \etal \cite{Chen2018SIG} model the appearance of a triangular mesh based on the UV coordinates and the viewing direction.
In contrast to traditional surface light fields we do not assume fixed illumination but instead condition on the light source.
This allows our model to manipulate illumination at inference time.
Furthermore, while all aforementioned models consider a single object, our approach is able to predict light fields for novel unseen objects.

\boldparagraph{Implicit 3D Representations}
Recently, implicit 3D geometry models have gained popularity \cite{Mescheder2019CVPR, Park2019CVPR, Chen2019CVPR} as they circumvent the discretization inherent to traditional explicit representations.
Implicit appearance models represent texture information by learning the mapping from 3D location to color \cite{Oechsle2019ICCV,Niemeyer2019ARXIV,Sitzmann2019NIPS,Saito2019ARXIV}.
However, all existing implicit representations are restricted to simple textures which represent only diffuse material properties.
In contrast, our model represents both diffuse and specular material properties as well as shadows.
Furthermore, modeling surface properties allows for capturing high-frequency details which cannot be captured with texture models (see \figref{Fig:chair} in our experiments).

\section{Method}
\label{Sec:Method}

\begin{figure}[t!]
	\includegraphics[width=\linewidth]{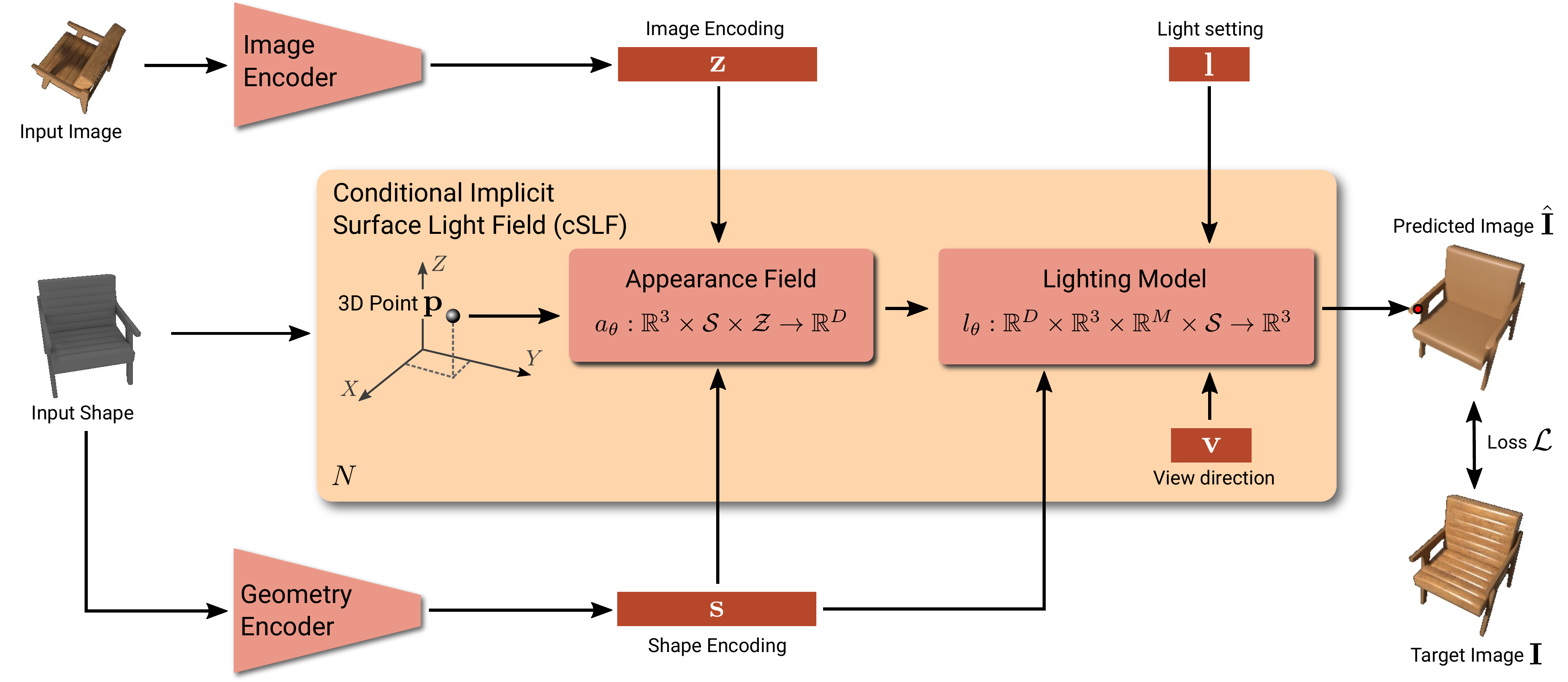}
	\vspace{-0.5cm}
	\caption{\textbf{Model Overview.} Our Conditional Implicit Surface Light Field (cSLF) takes an encoding of a real image ($\bz$) and corresponding 3D shape ($\bs$) as well as a light configuration ($\bl$) as input and outputs a color value for every 3D location $\bp\in\nR^3$.}
	\label{fig:model_overview}
	\vspace{-0.3cm}
\end{figure}

An overview of our model is provided in \figref{fig:model_overview}.
We take a 2D image and the corresponding 3D shape as input, and encode both into latent representations ($\bz$ and $\bs$) which are fed into our conditional implicit Surface Light Field (cSLF). Given a 3D point $\bp$, its viewing direction $\bv$ and the light configuration $\bl$, our cSLF model then outputs the predicted color value for this point.

We first describe the physical image formation process and traditional surface light field representations. Next, we introduce our cSLF model. Finally, we provide details about the network architectures and the loss functions.

\subsection{Background}
\textbf{Physically-based rendering} techniques \cite{Jakob2010} utilize the rendering equation to calculate the amount of light that radiates from a specific surface location into the direction of the camera.
Let $\bp\in\nR^3$ denote a 3D surface point and $\bv \in \nR^3$ the viewpoint of the camera, \ie, the vector from $\bp$ to the camera center. Let further $\br \in \nR^3$ denote the incoming light direction. The rendering equation describes how much of the light arriving at $\bp$ is reflected into the camera direction $\bv$:
\begin{equation}
L(\bp,\bv) =
\int_{\Omega} \text{svBRDF}(\bp,\br,\bv)
\cdot L_{\text{in}}(\mathbf{r})
\cdot (\bn^T\br) ~ d \br
\label{eq:renderingeq}
\end{equation}
Here, $\text{svBRDF}(\bp,\br,\bv)$ denotes the spatially varying bidirectional reflectance distribution function which models the interaction between light and the surface of a 3D object by determining the proportion of light reflected in outgoing direction $\bv$ from incoming light direction $\br$.
Furthermore, $L_{\text{in}}$ is the incoming radiance and $\bn\in\nR^3$ describes the surface normal at point $\bp$.

\vspace{0.1cm}

\noindent \textbf{Surface light fields} \cite{Gortler1996SIG, Wood2000SIG} instead directly represent the radiated light as a function that associates a color value $\bc\in\nR^3$ with every surface location $\mathbf{p}$ and view direction $\mathbf{v}$. Formally, they are described by the following mapping:
\begin{equation}
    L_\text{SLF}(\bp,\bv) : \nR^3 \times \nR^3 \rightarrow \nR^3
    \label{eq:SLF}
\end{equation}
While this mapping can be learned using neural networks \cite{Maximov2019ICCV,Chen2018SIG}, surface light fields do not allow for varying the light configuration. Instead, the illumination properties are ``hard-coded'' in this representation. In contrast, the model proposed in the following is conditioned on the light configuration, thus allowing the modification of illumination parameters at inference time.

\subsection{Conditional Implicit Surface Light Fields}
\label{sec:cSLF}

In this work, we aim at representing illumination-dependent visual appearance by learning a surface light field conditioned on a single light source. 
We refer to this representation as conditional implicit Surface Light Field (cSLF)
\begin{equation}
L_\text{cSLF}(\bp,\bv,\bl, \bs, \bz) : \nR^3 \times \nR^3  \times \nR^M \times \cS  \times \cZ \rightarrow \nR^3
\label{eq:cSLF}
\end{equation}
where $\bl \in \nR^M$ denotes the parameters of a point light source\footnote{In practice, we use a small area light source to avoid hard non-differentiable shadows.} (\eg, its location in 3D space and color). $\bs\in\cS$ and $\bz\in\cZ$ encode the object shape and image content, respectively.
We represent $L_\text{cSLF}$ implicitly using a neural network. 
In contrast to traditional surface light fields \eqref{eq:SLF}, our implicit representation allows for querying the light field for any light configuration $\bl$ and object encoding $(\bs, \bz)$.
This allows our model to generalize to novel, previously unseen 3D objects.

While our cSLF model considers a single point light source during training, more complex light settings, including spatial varying illumination, can be achieved by combining multiple evaluations of \eqnref{eq:cSLF} at inference time. Discretizing the domain $\Omega$ in \eqnref{eq:renderingeq} and letting $\cL=\{\bl_1,\dots,\bl_K\}$ denote a sampling-based approximation of the environment map (\ie, $\bl_k$ in $\cL$ corresponds to one pixel in the environment map), we obtain the aggregate surface light field as:
\begin{equation}
L(\bp,\bv,\cL, \bs, \bz) = \frac{1}{|\cL|}\sum_{\bl \in \cL} L_\text{cSLF}(\bp,\bv,\bl, \bs, \bz)
\label{eq:composite}
\end{equation}
In our experiments, we evaluate two different parameterizations of \eqnref{eq:cSLF}.

\boldparagraph{1-Step Model}
Our first model represents \eqref{eq:cSLF} using a single neural network:
\begin{equation}
    f_\theta(\bp,\bv,\bl, \bs, \bz) : \nR^3 \times \nR^3  \times \nR^M \times \cS  \times \cZ \rightarrow \nR^3
\label{eq:cSLF_single}
\end{equation}

\boldparagraph{2-Step Model}
Second, we split the parameterization of \eqnref{eq:cSLF} into a two-step process. 
We first map the 3D location $\bp$ conditioned on global shape ($\bs$) and image ($\bz$) information into an $D$-dimensional appearance feature $\bff$. The resulting \textit{appearance field} is described by:
\begin{equation}
	a_\theta (\bp, \bs, \bz) : \nR^3 \times \cS  \times \cZ \rightarrow \nR^D
\end{equation}
Note that in contrast to the global image feature, $\bff=a_\theta (\bp, \bs, \bz) $ is a localized appearance representation of the 3D object.
As $\bff$ is independent of the viewpoint $\bv$ and lighting $\bl$, it captures mostly material properties.
The interaction between the object surface and light is modeled using a \textit{lighting model} which
maps the appearance $\bff$, the viewpoint $\bv$, the light $\bl$ and the shape $\bs$ into an RGB value:
\begin{equation}
    l_\theta (\bff,\bv,\bl,\bs) : \nR^D \times \nR^3 \times \nR^M  \times \cS \rightarrow \nR^3
\end{equation}
Splitting the inference process into two steps allows for pre-computing the appearance features. Thus, when manipulating the light setting, only the lighting model $l_\theta (\bv,\bff,\bl,\bs)$ needs to be evaluated. This is particularly important when considering more complex light settings (\eg, environment lighting) as in \eqnref{eq:composite}.

\subsection{Network Architectures}

We now briefly discuss the network architectures. See supplementary for details.

\boldparagraph{Image Encoder}
We use the image encoder of \cite{Mescheder2019CVPR} which consists of a ResNet-18 that outputs a 512-dimensional image encoding $\bz$.

\boldparagraph{Geometry Encoder}
To provide global shape information to our model, we sample 2048 points from the object surface and use the PointNet-based residual network of \cite{Mescheder2019CVPR} which encodes the 3D point cloud into a shape vector $\bs$.

\boldparagraph{Surface Light Field}
We parameterize the networks $f_\theta(\cdot)$, $a_\theta(\cdot)$ and $l_\theta(\cdot)$ of our surface light field model using fully-connected residual networks.

\subsection{Loss Functions}

We train our model in a supervised fashion on photo-realistic renderings of 3D models.
More specifically, we randomly sample a batch of 3D models along with one random input and one random target view. We also randomly sample the point light source location and color ($\bl$).
We randomly select $N$ pixels from the target view and project them into 3D, resulting in a set of 3D points $\bp$.
We also render photo-realistic RGB images for both the input and the target view using the Blender Cycles renderer.
During training, we apply a photometric loss between the predicted color values $\mathbf{\hat{I}}$ and the ground truth color values $\mathbf{I}$:
\begin{equation}\label{eq:loss}
	\mathcal{L}(\mathbf{\hat{I}}, \mathbf{I}) = \frac{1}{|\cB|} \sum_{b\in\cB} {\Vert \mathbf{\hat{I}}_b - \mathbf{I}_b\Vert}_{1}
\end{equation}
where $\hat{\bI}_b$ represents the predicted color values for all $N$ points of batch $\cB$.

To evaluate the generative capabilities of our surface light field representation, we auto-encode the image $\bI$ using a Variational Autoencoder (VAE) \cite{Kingma2014ICLR,Rezende2014ICML}. 
More specifically, we use a ResNet-based encoder $q_\phi(\bz|\bI,\bs)$ which embeds an image $\bI$ and a shape representation $\bs$ into a latent code $\bz$.
The decoder is given by the surface light field model described in \secref{sec:cSLF} which predicts a color value for every pixel, given its 3D location $\bp$, the viewpoint $\bv$, the light configuration $\bl$, the object shape $\bs$ and the latent code $\bz$.
The loss function for training our VAE  model is thus given by
\begin{align}
\cL_{\text{VAE}} = \frac{1}{|\cB|} \sum_{b\in\cB} \left[ \beta \; \text{KL}(q_\phi(\bz_b|\bI_b,\bs_b) \parallel \cN(\bz_b|0,\bI))
 +  {\Vert \mathbf{\hat{I}}_b - \bI_b\Vert}_{1} \right]
\label{eq:vae-objective}
\end{align}
where $\text{KL}(\cdot)$ represents the Kullback-Leibler divergence, $\bz_b$ follows the posterior distribution $q_\phi(\bz|\bI_b,\bs_b) $
and $\beta$ is a trade-off parameter \cite{Higgins2017ICLR} between the KL-divergence and the reconstruction loss.

\section{Experimental Evaluation}
\label{sec:Experiments}

We conduct three different types of experiments.
First, we investigate the ability of our method to capture the surface light field of a single object illuminated by a point light source.
Next, we %
use our model to predict the surface light field
of previously unseen objects from a single image. %
Furthermore, we provide results for geometry reconstruction and appearance prediction from a single image by applying our method to the output of
Occupancy Network \cite{Mescheder2019CVPR} reconstructions.
Finally, we %
show qualitative results from our generative model which learns the conditional distribution of surface light fields given the object geometry.

\boldparagraph{Dataset}
While ShapeNet~\cite{Chang2015ARXIV} provides a large variety of 3D models, these models are not suitable for our experiments as they lack realistic materials.
We therefore use the Photoshapes~dataset~\cite{Park2018SIG} which contains $5452$ meshes of chairs with a large variety of realistic materials.
We split the dataset into a training set, a validation set and a test set with ratio $7:1:2$.
In our single object experiment, we further show results on a high-quality 3D car model from \url{free3d.com}.
For testing our models on real-world data we use the Pix3D~dataset~\cite{SunPix3D2018CVPR}.

\boldparagraph{Rendering}
We use the Blender Cycles Renderer \cite{Blender} to generate ground truth RGB images and depth maps from random viewpoints and light locations,
sampled from the upper hemisphere.
More details on the rendering process are provided in the supplementary.

\subsection{Single Object Experiment}
\begin{figure}[h]
    \centering
    \includegraphics[width=\linewidth]{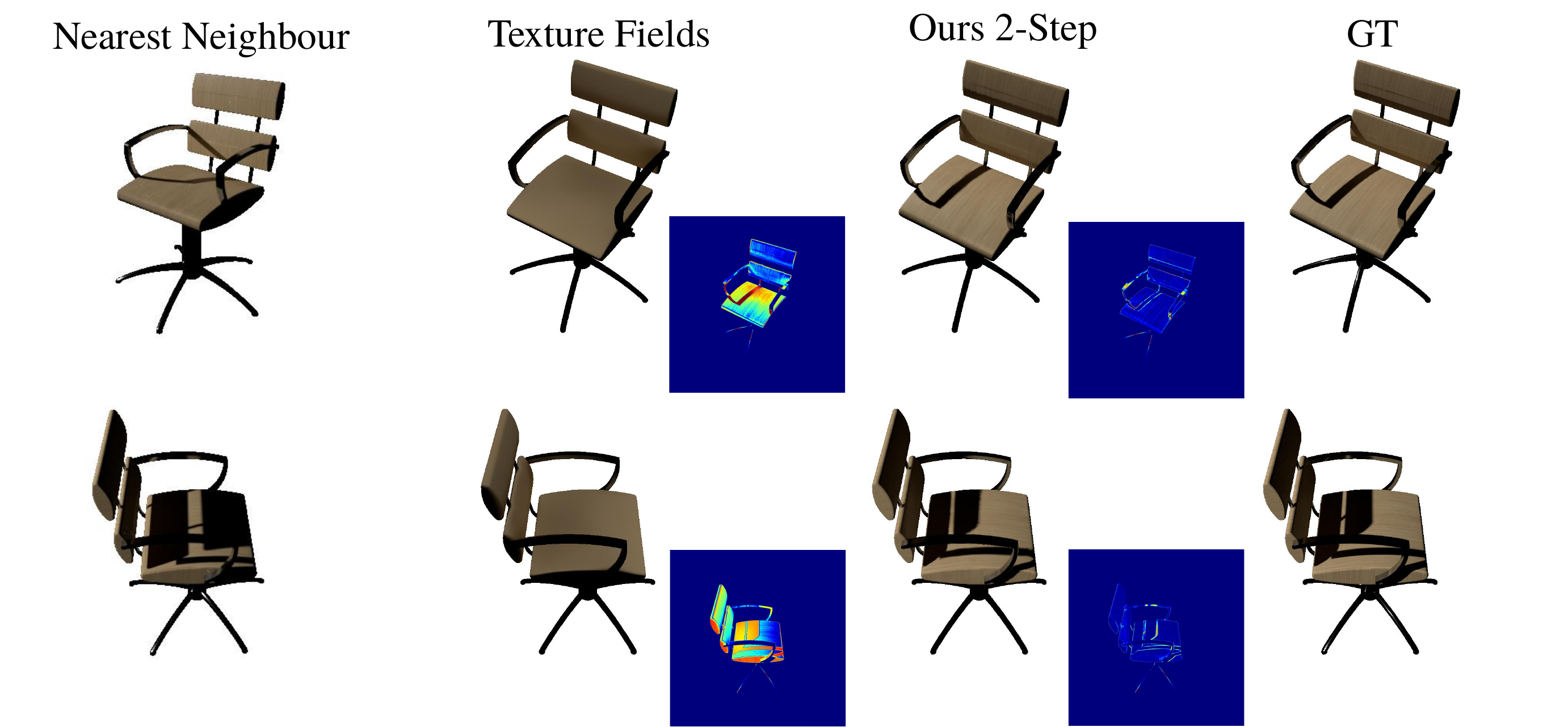} 
    \caption{\textbf{Shadow Analysis.} Comparison of TextureFields~\cite{Oechsle2019ICCV} and our 2-step model with respect to the nearest neighbor and ground truth for two different views. The small inlet shows the reconstruction error (blue=low error, red=large error).
	}
    \label{Fig:chair}
\end{figure}

\begin{figure}[h]
    \centering
    \includegraphics[width=\linewidth]{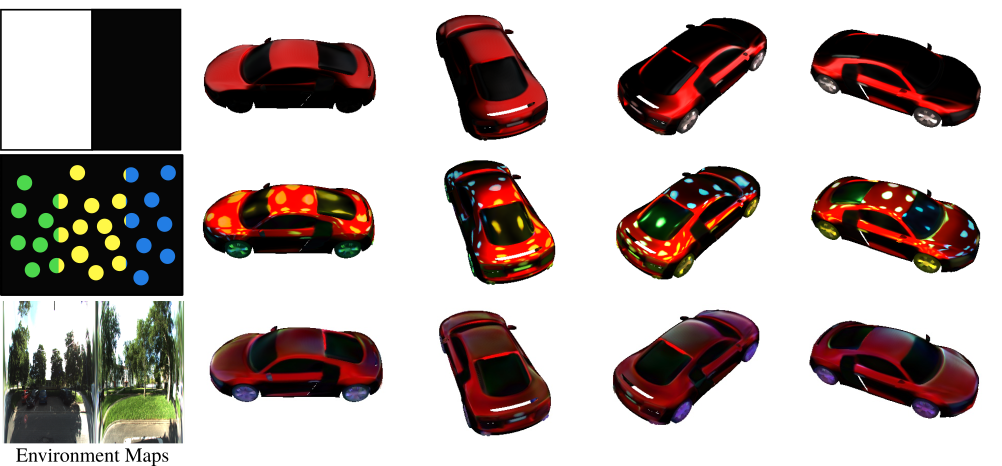}
    \caption{\textbf{Reflection Analysis.} We show results for a highly specular car with three different environment maps.
    }
    \label{Fig:car}
\end{figure}

We first investigate the representation power of our model \wrt texture, reflections and shadows by training it on a single object and evaluating it on novel unseen viewpoints and light conditions.
To analyze shadows, we choose a chair with thin armrests in \figref{Fig:chair}.
For analyzing reflections, we use a car model which contains highly specular materials in \figref{Fig:car}.
For training, we render 50 views with 30 lighting conditions each.
For evaluation, we choose views that are substantially different from the training views \wrt viewpoint and light location. We also show the nearest neighbor from the training set (\figref{Fig:chair} left).

\boldparagraph{Shadow Analysis}
\figref{Fig:chair} shows a comparison of our model to Texture Fields \cite{Oechsle2019ICCV}, a state-of-the-art implict model for texture representation.
As illustrated in \figref{Fig:chair}, our method successfully infers shadows and textural details for novel view and lighting configurations.
Note that synthesizing accurate shadows requires reasoning about the underlying geometry and is hence very challenging.
In contrast, the texture-based model \cite{Oechsle2019ICCV} ``averages'' shadows and texture details due to its inability to capture illumination effects.

\boldparagraph{Reflection Analysis}
To further investigate the capability of our model to accurately represent surface light fields, we apply our method to a car model which consists of highly specular materials in \figref{Fig:car}.
By representing complex illumination as a combination of light sources, our method is able to reason about the appearance of 3D objects in complex illumination settings.
We represent environment maps as the superposition of point light sources using \eqnref{eq:composite} and show examples of the car model rendered using different environment maps in \figref{Fig:car}.
We find that our model is able to accurately capture specular reflections, which allows for rendering 3D objects in complex illumination environments.

\subsection{Single View Appearance Prediction}
While reasoning about the inter-dependence between light, surface, and viewpoint of a single object is a challenging task, we now address the even more challenging problem of inferring surface light fields for novel, previously unseen objects.
To successfully solve this task, the network must implicitly infer appearance properties and geometric properties (\eg, surface normals, shadowing effects, \etc) of objects from the input image and shape encoding.
For this experiment, we render images and depth maps from 10 different viewpoints per object with 4 randomly sampled light locations per view.
We train our model using an image resolution of $256^2$ pixels.

\boldparagraph{Baselines}
As many existing neural rendering techniques use 2D CNNs for image synthesis, we consider a CNN-based image-to-image translation network as a baseline for our method.
Similar to~\cite{Oechsle2019ICCV, Thies2019SIG}, we employ a U-Net architecture \cite{Ronneberger2015MICCAI}. %
which maps a depth image from the same viewpoint to an RGB image, conditioned on the light location and the geometry of the object.
We refer to this baseline as Img2Img, see supplementary for details.
As a second baseline, we use Texture Fields \cite{Oechsle2019ICCV}, a state-of-the-art implicit texture model. 

\boldparagraph{Metrics}
We evaluate all methods on three different metrics.
First, we compute the FID score between a set of predicted images and ground truth images \cite{Heusel2017NIPS}.
We further report the Structure Similarity Image Metric (SSIM) \cite{Wang2004TIP} and 
a $l_1$ distance measure in the Inception feature space between the predicted image and the ground truth as in \cite{Oechsle2019ICCV}.

\begin{figure}[t]
    \centering
    \begin{tabular}{c@{}c@{}c@{}c@{}c@{}c}
	Input & TexFields & Img2Img & Ours 1-s & Ours 2-s & GT\\
	\includegraphics[width=0.16\linewidth,trim={.75cm .75cm .75cm .75cm},clip]{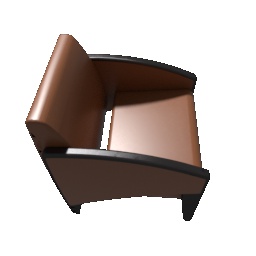} &
	\includegraphics[width=0.16\linewidth,trim={.75cm .75cm .75cm .75cm},clip]{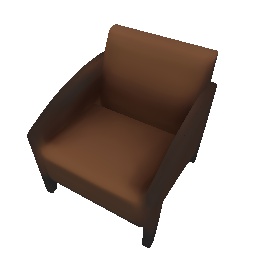} &
	\includegraphics[width=0.16\linewidth,trim={.75cm .75cm .75cm .75cm},clip]{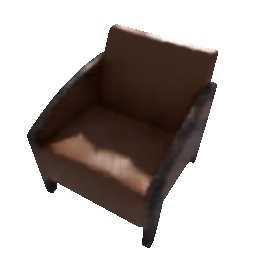} &
	\includegraphics[width=0.16\linewidth,trim={.75cm .75cm .75cm .75cm},clip]{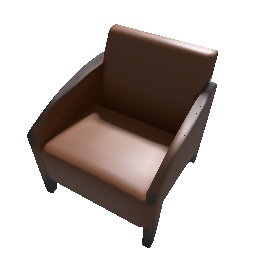} &
	\includegraphics[width=0.16\linewidth,trim={.75cm .75cm .75cm .75cm},clip]{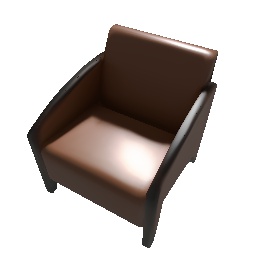} &
	\includegraphics[width=0.16\linewidth,trim={.75cm .75cm .75cm .75cm},clip]{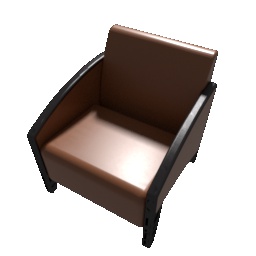} \\
	\includegraphics[width=0.16\linewidth,trim={.75cm .75cm .75cm .75cm},clip]{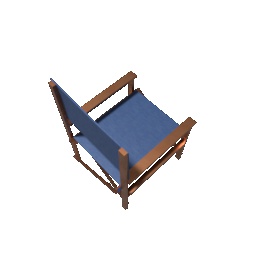} &
	\includegraphics[width=0.16\linewidth,trim={.75cm .75cm .75cm .75cm},clip]{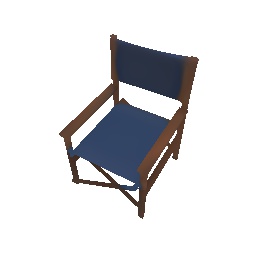} &
	\includegraphics[width=0.16\linewidth,trim={.75cm .75cm .75cm .75cm},clip]{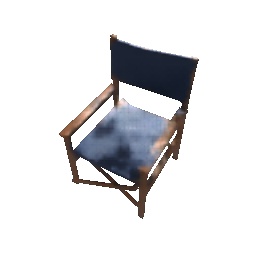} &
	\includegraphics[width=0.16\linewidth,trim={.75cm .75cm .75cm .75cm},clip]{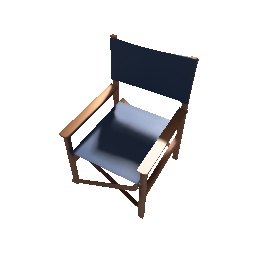} &
	\includegraphics[width=0.16\linewidth,trim={.75cm .75cm .75cm .75cm},clip]{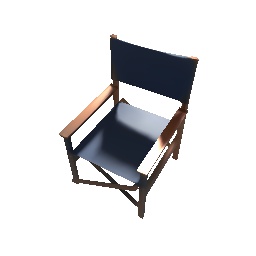} &
	\includegraphics[width=0.16\linewidth,trim={.75cm .75cm .75cm .75cm},clip]{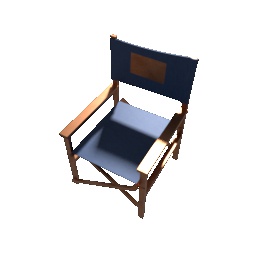} \\
	\includegraphics[width=0.16\linewidth,trim={.75cm 1.75cm .75cm 1.75cm},clip]{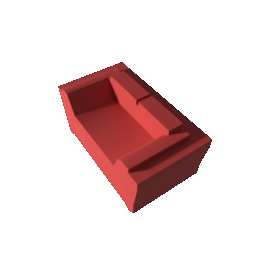} &
	\includegraphics[width=0.16\linewidth,trim={.75cm 1.75cm .75cm 1.75cm},clip]{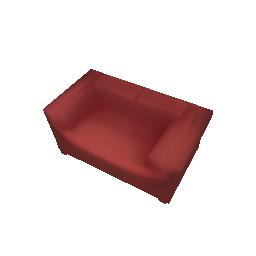} &
	\includegraphics[width=0.16\linewidth,trim={.75cm 1.75cm .75cm 1.75cm},clip]{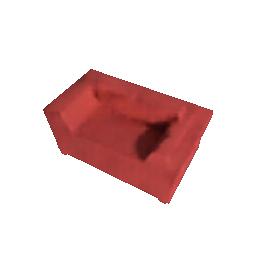} &
	\includegraphics[width=0.16\linewidth,trim={.75cm 1.75cm .75cm 1.75cm},clip]{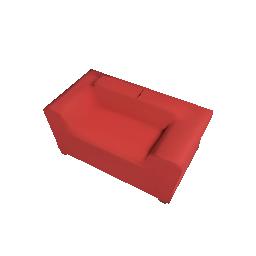} &
	\includegraphics[width=0.16\linewidth,trim={.75cm 1.75cm .75cm 1.75cm},clip]{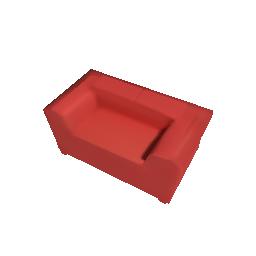} &
	\includegraphics[width=0.16\linewidth,trim={.75cm 1.75cm .75cm 1.75cm},clip]{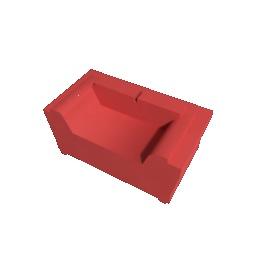} \\
	\includegraphics[width=0.16\linewidth,trim={.75cm 1.75cm .75cm .75cm},clip]{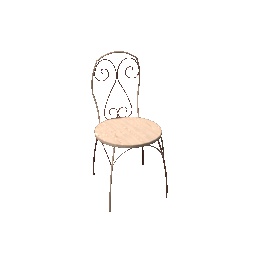} &
	\includegraphics[width=0.16\linewidth,trim={.75cm 1.75cm .75cm .75cm},clip]{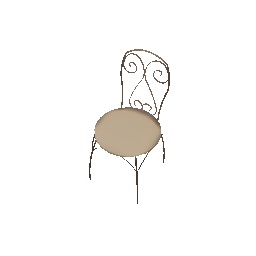} &
	\includegraphics[width=0.16\linewidth,trim={.75cm 1.75cm .75cm .75cm},clip]{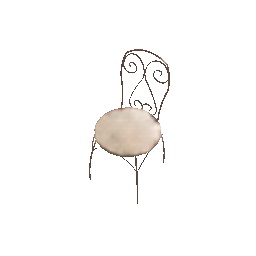} &
	\includegraphics[width=0.16\linewidth,trim={.75cm 1.75cm .75cm .75cm},clip]{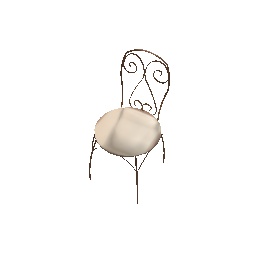} &
	\includegraphics[width=0.16\linewidth,trim={.75cm 1.75cm .75cm .75cm},clip]{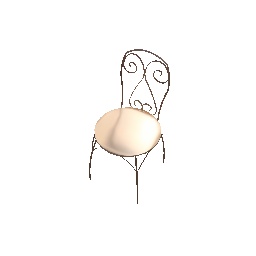} &
	\includegraphics[width=0.16\linewidth,trim={.75cm 1.75cm .75cm .75cm},clip]{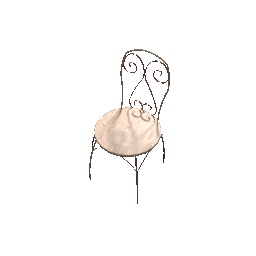} \\
\end{tabular}

    \caption{\textbf{Single View Appearance Prediction.} 
    Results of our method and the provided baselines for predicting the appearance of a novel, previously unseen object.
     }
    \label{fig:singleview_02728}
\end{figure}

\boldparagraph{Surface Light Field Prediction}
In our first experiment, we predict the appearance of a novel object from a single RGB image and the corresponding ground truth geometry embedding.
\figref{fig:singleview_02728} shows the predictions for different light locations using our method, the Img2Img baseline and Texture Fields~\cite{Oechsle2019ICCV}.

\vspace{0.2cm}
\noindent\emph{Does our model predict accurate textures?}
\figref{fig:singleview_02728} shows that our method is able to reconstruct details from the input view like the brown armrests and chair legs. 
While our methods reconstruct plausible colors, the Img2Img baseline produces artifacts at borders between two color regions.

\vspace{0.2cm}
\noindent\emph{Does our model predict plausible shadows for different light configurations?}
\figref{fig:singleview_02728} shows that our method is able to reason about shadows. Note that even for the geometry of the last example, our model predicts a rough silhouette.
The Img2Img baseline, in contrast, results in very noisy predictions.
The Texture Field baseline lacks a physical understanding of image formation and thus ``averages'' shadows in the training data, leading to blurry results.

\vspace{0.2cm}
\noindent\emph{Does our model represent specular reflections?}
The examples in \figref{fig:singleview_02728} contain specularities as well as more diffuse reflection properties.
In contrast to the Img2Img Baseline and Texture Fields, our method is able to predict plausible specular reflections (brown chair) as well as diffuse materials (red couch).

\boldparagraph{Quantitative Comparison}
In \tabref{tab:baselines}, we quantitatively compare our method to the baselines. While our method and the Img2Img baseline compare similarly in SSIM, our method performs favorably in terms of FID and Feature-$l_1$ distance.
As expected, the Texture Field baseline is significantly worse compared to the other methods, since it does not capture the full visual appearance. 
However, we remark that none of the metrics focuses explicitly on shadows or surface properties, highlighting the importance of our qualitative comparisons.

\begin{table}[t]
    \centering
    \setlength{\tabcolsep}{0.5cm}
    \begin{tabular}{lccc}
\toprule
{} &              FID &            SSIM & Feature-$\ell_1$ \\
\midrule
Texture Fields  &           33.46 &           0.94 &            0.16 \\
Img2Img Translation &           27.01 &  \textbf{0.96} &            0.14 \\
Ours 1-step     &           21.26 &           \textbf{0.96} &   \textbf{0.13} \\
Ours 2-step     &  \textbf{21.20} &           \textbf{0.96} &   \textbf{0.13} \\
\bottomrule
\end{tabular}

    \vspace{0.3cm}
    \caption{\textbf{Single View Appearance Prediction.} Quantitative comparison of our approach with the Img2Img baseline. 
    Our method performs better on the FID and Feature-$l_1$ metric. In the less accurate SSIM metric all methods perform similarly.}
	\label{tab:baselines} 
\end{table}

\begin{table}[t]
    \centering
    \setlength{\tabcolsep}{0.5cm}
    \begin{tabular}{lccc}
\toprule
{} &              FID &            SSIM & Feature-$\ell_1$ \\
\midrule
Ours 1-step (2 views)     &           24.51 &           0.95 &            0.14 \\
Ours 2-step (2 views)     &           24.72 &           0.95 &            0.14 \\
Ours 2-step (w/o shape $\mathbf{s}$) &           24.30 &           0.95 &            0.14 \\
Ours 2-step           &  \textbf{21.20} &  \textbf{0.96} &   \textbf{0.13} \\
\bottomrule
\end{tabular}

    \vspace{0.3cm}
    \caption{\textbf{Ablation Study.} Quantitative results when removing the input shape encoding ($\bs$) and reducing the number of training views per object to 2.}
    \label{tab:ablation} 
\end{table}

\boldparagraph{Ablation Study}
We hypothesized that our model relies on geometric information (shape encoding $\bs$ in \figref{fig:model_overview}) to model shadows and surface properties.
We therefore compare our approach against a version without shape encoding (Ours w/o shape $\bs$) in  
\tabref{tab:ablation}.
Indeed, removing the shape encoding leads to a degradation in all metrics.
We further investigate if our method can also be trained on only two views per object with different light settings (2 views).
While two input views per object are sufficient for learning to infer plausible appearance from a single image, the complex physical light formation process cannot be modeled well. We provide qualitative results for this study in the supplementary.

\begin{figure}
    \begin{tabular}{cc@{}c@{}c@{}c}
	\includegraphics[width=0.19\linewidth,trim={.75cm .5cm .75cm .5cm},clip]{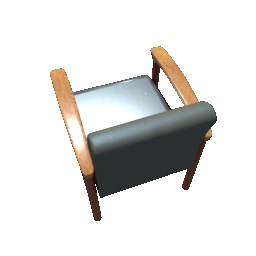} &
	\includegraphics[width=0.19\linewidth,trim={0cm 0cm 0cm 0cm},clip]{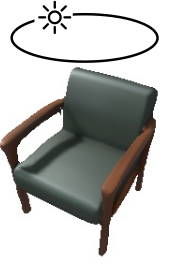} &
	\includegraphics[width=0.19\linewidth,trim={0cm, 0cm 0cm 0cm},clip]{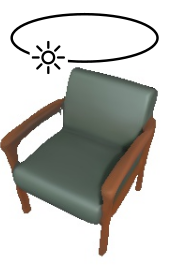} &
	\includegraphics[width=0.19\linewidth,trim={0cm 0cm 0cm 0cm},clip]{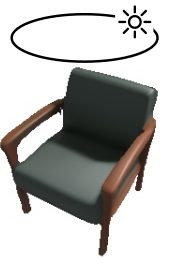} &
	\includegraphics[width=0.19\linewidth,trim={.0cm 0cm 0cm 0cm},clip]{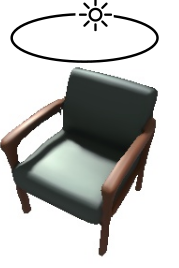} \\
	\includegraphics[width=0.19\linewidth,trim={.75cm .5cm .75cm .cm},clip]{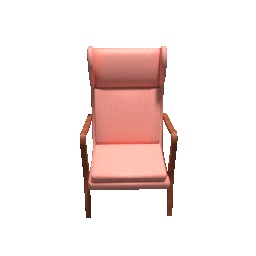} &
	\includegraphics[width=0.19\linewidth,trim={.75cm .5cm .75cm .5cm},clip]{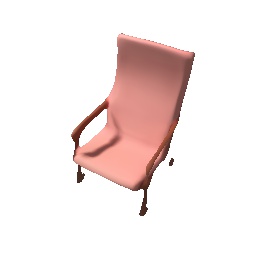} &
	\includegraphics[width=0.19\linewidth,trim={.75cm .5cm .75cm .5cm},clip]{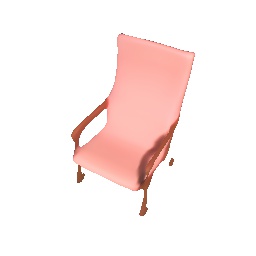} &
	\includegraphics[width=0.19\linewidth,trim={.75cm .5cm .75cm .5cm},clip]{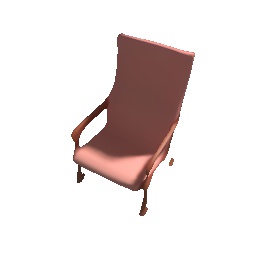} &
	\includegraphics[width=0.19\linewidth,trim={.75cm .5cm .75cm .5cm},clip]{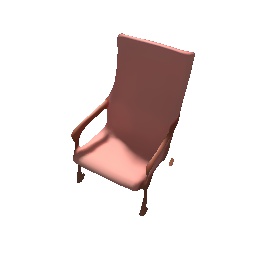} \\
	\includegraphics[width=0.19\linewidth,trim={.75cm .5cm .75cm .5cm},clip]{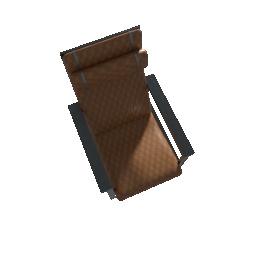} &
	\includegraphics[width=0.19\linewidth,trim={.75cm .5cm .75cm .5cm},clip]{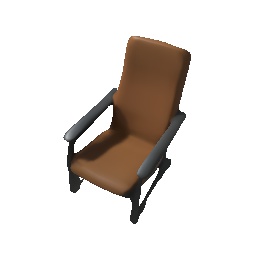} &
	\includegraphics[width=0.19\linewidth,trim={.75cm .5cm .75cm .5cm},clip]{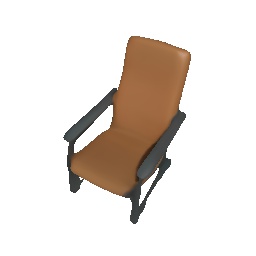} &
	\includegraphics[width=0.19\linewidth,trim={.75cm .5cm .75cm .5cm},clip]{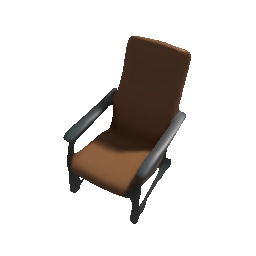} &
	\includegraphics[width=0.19\linewidth,trim={.75cm .5cm .75cm .5cm},clip]{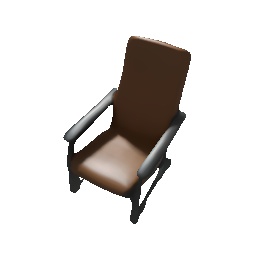} \\
	\includegraphics[width=0.19\linewidth,trim={.75cm .5cm .75cm .5cm},clip]{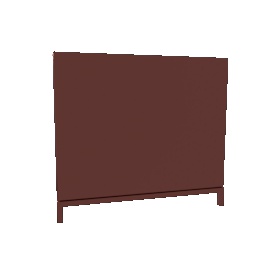} &
	\includegraphics[width=0.19\linewidth,trim={.75cm .5cm .75cm .5cm},clip]{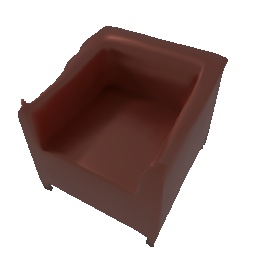} &
	\includegraphics[width=0.19\linewidth,trim={.75cm .5cm .75cm .5cm},clip]{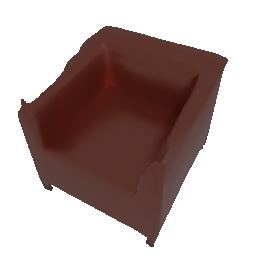} &
	\includegraphics[width=0.19\linewidth,trim={.75cm .5cm .75cm .5cm},clip]{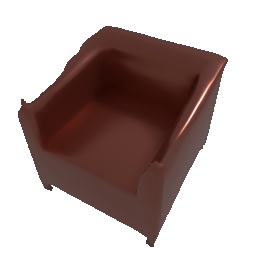} &
	\includegraphics[width=0.19\linewidth,trim={.75cm .5cm .75cm .5cm},clip]{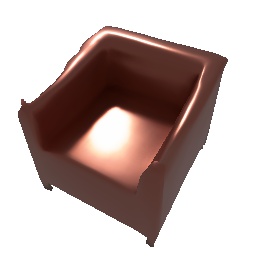} \\
	\includegraphics[width=0.19\linewidth,trim={.75cm .5cm .75cm .5cm},clip]{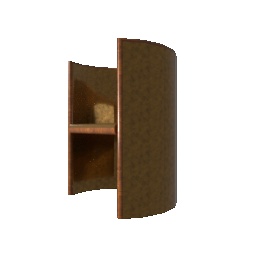} &
	\includegraphics[width=0.19\linewidth,trim={.75cm .5cm .75cm .5cm},clip]{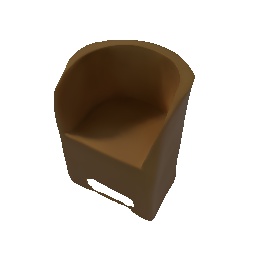} &
	\includegraphics[width=0.19\linewidth,trim={.75cm .5cm .75cm .5cm},clip]{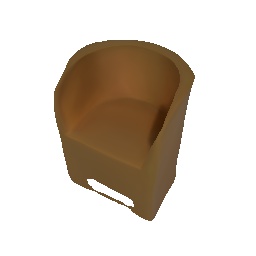} &
	\includegraphics[width=0.19\linewidth,trim={.75cm .5cm .75cm .5cm},clip]{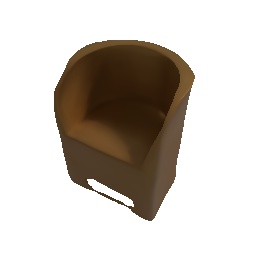} &
	\includegraphics[width=0.19\linewidth,trim={.75cm .5cm .75cm .5cm},clip]{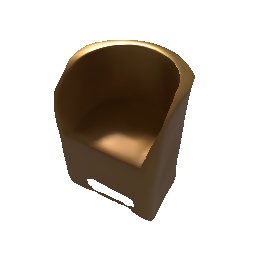} \\
	\includegraphics[width=0.19\linewidth,trim={.75cm .5cm .75cm .5cm},clip]{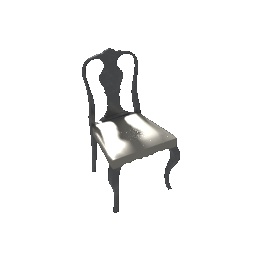} &
	\includegraphics[width=0.19\linewidth,trim={.75cm .5cm .75cm .5cm},clip]{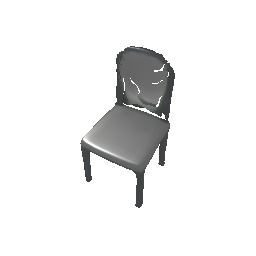} &
	\includegraphics[width=0.19\linewidth,trim={.75cm .5cm .75cm .5cm},clip]{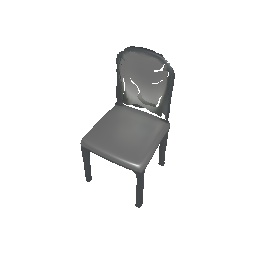} &
	\includegraphics[width=0.19\linewidth,trim={.75cm .5cm .75cm .5cm},clip]{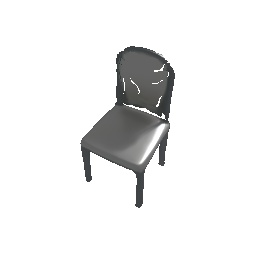} &
	\includegraphics[width=0.19\linewidth,trim={.75cm .5cm .75cm .5cm},clip]{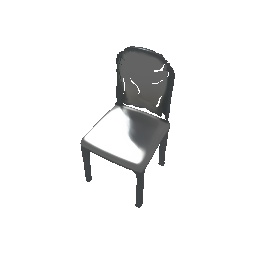} \\
	Input Image & \multicolumn{4}{c}{Predictions}
\end{tabular}
    \caption{\textbf{Image-based Reconstruction.} Reconstruction of geometry and appearance based on a single RGB image. We show the input image in the first column and predictions of our 2-step model for different light configurations in the following columns.
    \label{fig:dvr}
}
\end{figure}

\boldparagraph{Image-based Reconstruction}
While we use ground truth geometry as input in all previous experiments, such information is often not available in real-world settings.
We therefore investigate whether our approach allows for inferring the appearance when using inaccurate image-based 3D reconstructions as input.
Towards this goal, we train the model proposed in \cite{Niemeyer2019ARXIV}
on our renderings and combine it with our surface light field to reconstruct both geometry and appearance from a single RGB image.
As evidenced by \figref{fig:dvr}, our surface light field model allows for inferring accurate textures, reflections and shadow effects even when using less accurate, inferred 3D geometry as input.

\begin{figure}[h]
    \centering
    \begin{tabular}{c@{}c@{}c@{}c@{}c}
	\includegraphics[width=0.16\linewidth,trim={0.1cm .1cm .1cm .1cm},clip]{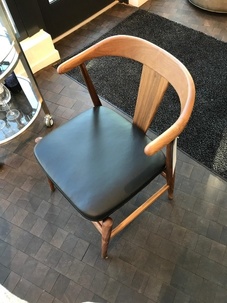} &
	\includegraphics[width=0.21\linewidth,trim={0.5cm 1cm 0.5cm 0cm},clip]{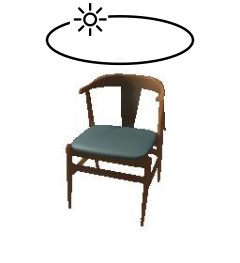} &
	\includegraphics[width=0.21\linewidth,trim={0.5cm 1cm 0.5cm 0cm},clip]{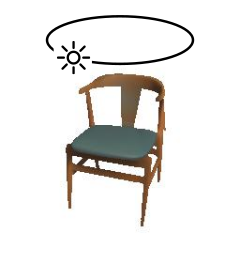} &
	\includegraphics[width=0.21\linewidth,trim={0.5cm 1cm 0.5cm 0cm},clip]{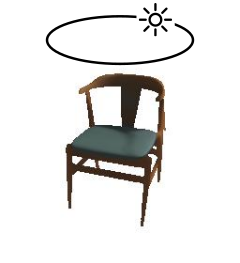} &
	\includegraphics[width=0.21\linewidth,trim={0.5cm 1cm 0.5cm 0cm},clip]{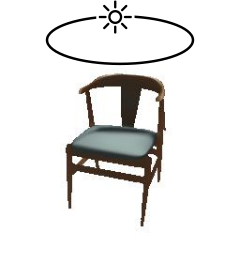} \\
	\includegraphics[width=0.16\linewidth,trim={.1cm .1cm .1cm .1cm},clip]{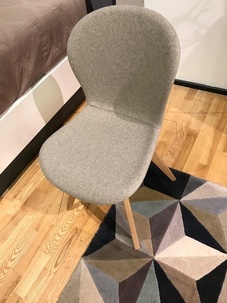} &
	\includegraphics[width=0.21\linewidth,trim={1.5cm 1.5cm 1.5cm 1.5cm},clip]{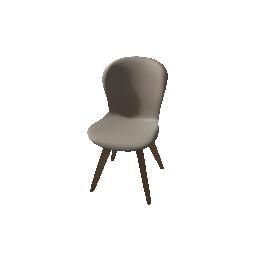} &
	\includegraphics[width=0.21\linewidth,trim={1.5cm 1.5cm 1.5cm 1.5cm},clip]{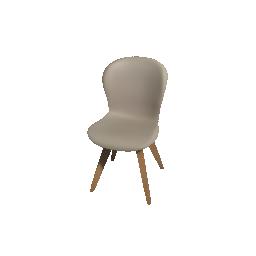} &
	\includegraphics[width=0.21\linewidth,trim={1.5cm 1.5cm 1.5cm 1.5cm},clip]{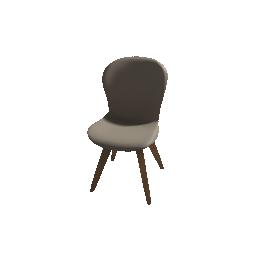} &
	\includegraphics[width=0.21\linewidth,trim={1.5cm 1.5cm 1.5cm 1.5cm},clip]{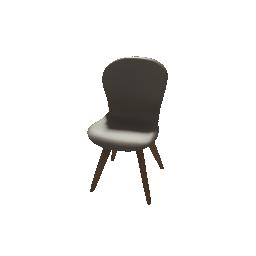} \\
	\includegraphics[width=0.16\linewidth,trim={.1cm .1cm .1cm .1cm},clip]{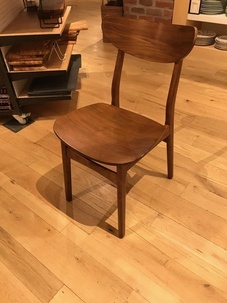} &
	\includegraphics[width=0.21\linewidth,trim={1.5cm 1.5cm 1.5cm 1.5cm},clip]{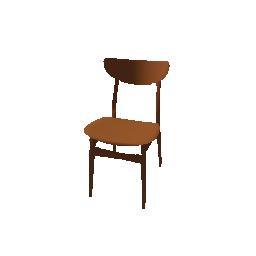} &
	\includegraphics[width=0.21\linewidth,trim={1.5cm 1.5cm 1.5cm 1.5cm},clip]{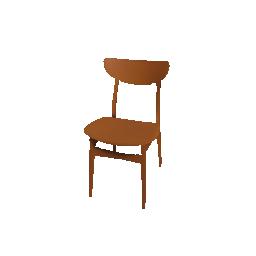} &
	\includegraphics[width=0.21\linewidth,trim={1.5cm 1.5cm 1.5cm 1.5cm},clip]{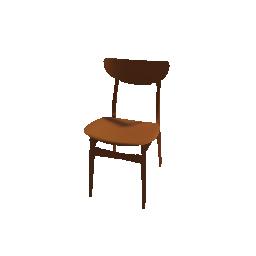} &
	\includegraphics[width=0.21\linewidth,trim={1.5cm 1.5cm 1.5cm 1.5cm},clip]{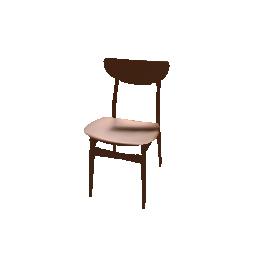} \\
	\includegraphics[width=0.16\linewidth,trim={.5cm .5cm .5cm .5cm},clip]{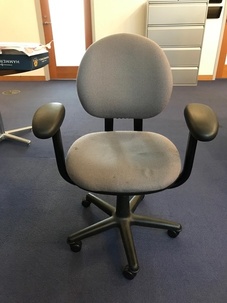} &
	\includegraphics[width=0.21\linewidth,trim={1.5cm 1.5cm 1.5cm 1.5cm},clip]{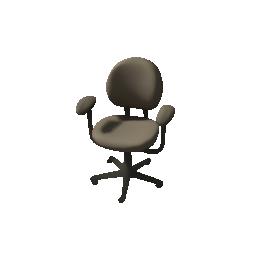} &
	\includegraphics[width=0.21\linewidth,trim={1.5cm 1.5cm 1.5cm 1.5cm},clip]{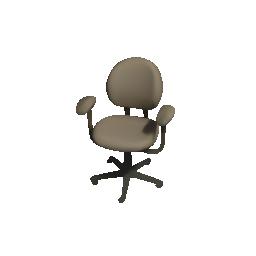} &
	\includegraphics[width=0.21\linewidth,trim={1.5cm 1.5cm 1.5cm 1.5cm},clip]{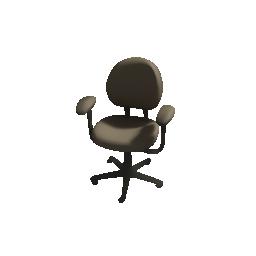} &
	\includegraphics[width=0.21\linewidth,trim={1.5cm 1.5cm 1.5cm 1.5cm},clip]{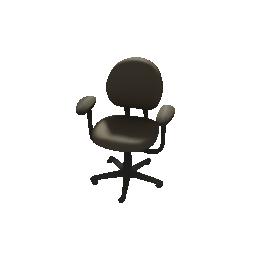} \\
\end{tabular}
 
    \caption{\textbf{Real Images.} Results of our method when using real images of unseen objects as input. 
    Even though our model is trained on synthetic images, it predicts plausible appearance for real images. 
    Note that we used masked images as input.}
    \label{fig:pix3d} 
\end{figure}

\boldparagraph{Real Images}
We now investigate if our approach trained on synthetic data generalizes to real images as input. 
We use the Pix3D~dataset~\cite{SunPix3D2018CVPR} for this purpose, as it offers real input images with aligned 3D geometry 
and apply the provided instance masks to the input images.
As illustrated in \figref{fig:pix3d}, our approach successfully reasons about the appearance of real objects despite being trained on synthetic data only (both in terms of input and ground truth supervision).

\subsection{Generative Model}

\begin{figure}[h]
    \centering
    \def\gfxwidthA{0.166}
\begin{tabular}{c@{}c@{}c@{}c@{}c@{}c@{}c@{}c}
	\includegraphics[width=\gfxwidthA\linewidth,trim={1.cm 1.cm 1.cm 1.cm},clip]{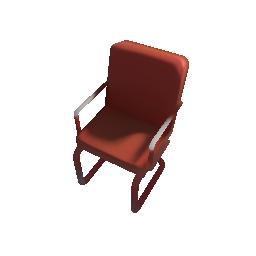} &
	\includegraphics[width=\gfxwidthA\linewidth,trim={1.cm 1.cm 1.cm 1.cm},clip]{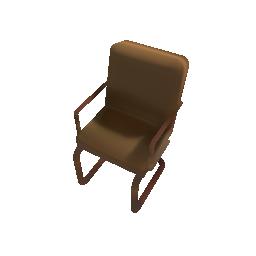} & 
	\includegraphics[width=\gfxwidthA\linewidth,trim={1.cm 1.cm 1.cm 1.cm},clip]{gfx/generative/samples/transparent_background/0001.jpg} &
	\includegraphics[width=\gfxwidthA\linewidth,trim={1.cm 1.cm 1.cm 1.cm},clip]{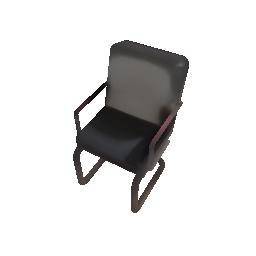} &
	\includegraphics[width=\gfxwidthA\linewidth,trim={1.cm 1.cm 1.cm 1.cm},clip]{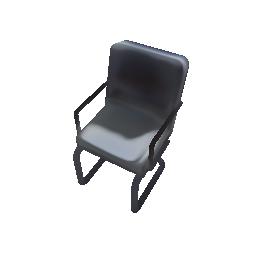}&
	\includegraphics[width=\gfxwidthA\linewidth,trim={1.cm 1.cm 1.cm 1.cm},clip]{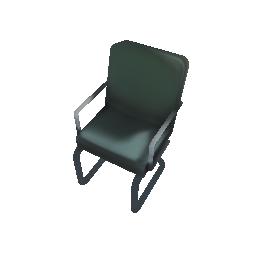}\\
	\multicolumn{6}{c}{Randomly sampled latent code $\bz$} \\
	\includegraphics[width=\gfxwidthA\linewidth,trim={1.cm 1.cm 1.cm 1.cm},clip]{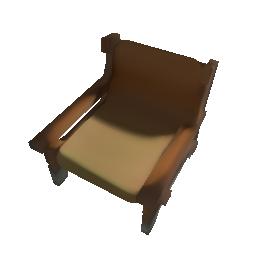} &
	\includegraphics[width=\gfxwidthA\linewidth,trim={1.cm 1.cm 1.cm 1.cm},clip]{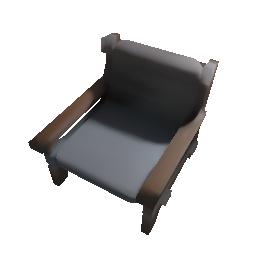} & 
	\includegraphics[width=\gfxwidthA\linewidth,trim={1.cm 1.cm 1.cm 1.cm},clip]{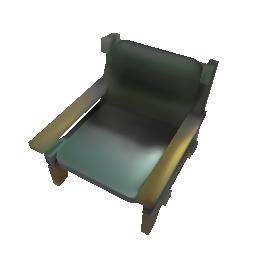} &
	\includegraphics[width=\gfxwidthA\linewidth,trim={1.cm 1.cm 1.cm 1.cm},clip]{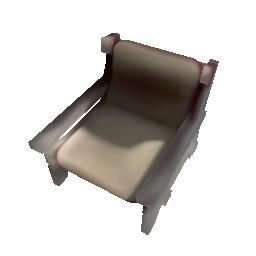} &
	\includegraphics[width=\gfxwidthA\linewidth,trim={1.cm 1.cm 1.cm 1.cm},clip]{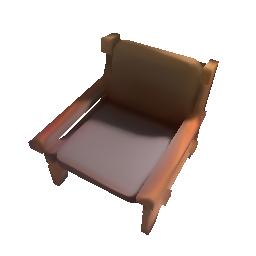}&
	\includegraphics[width=\gfxwidthA\linewidth,trim={1.cm 1.cm 1.cm 1.cm},clip]{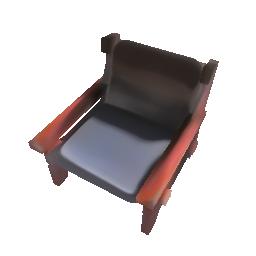}\\
	\multicolumn{6}{c}{Interpolated latent code $\bz$} \\
\end{tabular}

    \def\gfxwidthB{0.19}
\begin{tabular}{c@{}|c@{}c@{}c@{}c}	
	\includegraphics[width=\gfxwidthB\linewidth,trim={1.cm 1.cm 1.cm 1.cm},clip]{gfx/singleview/input/shape03629_white.jpg} &
	\includegraphics[width=\gfxwidthB\linewidth,trim={1.cm 1.cm 1.cm 1.cm},clip]{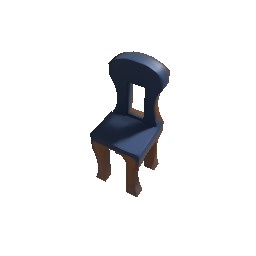} & 
	\includegraphics[width=\gfxwidthB\linewidth,trim={1.cm 1.cm 1.cm 1.cm},clip]{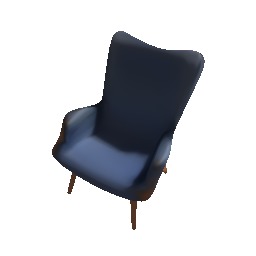} &
	\includegraphics[width=\gfxwidthB\linewidth,trim={1.cm 1.cm 1.cm 1.cm},clip]{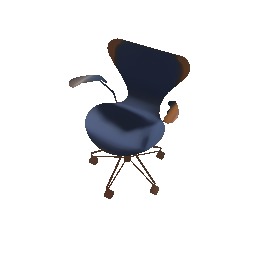} &
	\includegraphics[width=\gfxwidthB\linewidth,trim={1.cm 1.cm 1.cm 1.cm},clip]{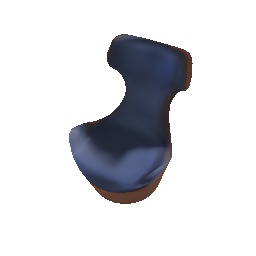}\\	
	\includegraphics[width=\gfxwidthB\linewidth,trim={01.cm 1.cm 1.cm 1.cm},clip]{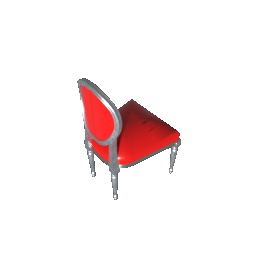} &
	\includegraphics[width=\gfxwidthB\linewidth,trim={1.cm 1.cm 1.cm 1.cm},clip]{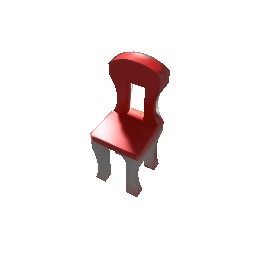} & 
	\includegraphics[width=\gfxwidthB\linewidth,trim={1.cm 1.cm 1.cm 1.cm},clip]{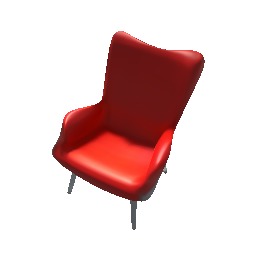} &
	\includegraphics[width=\gfxwidthB\linewidth,trim={1.cm 1.cm 1.cm 1.cm},clip]{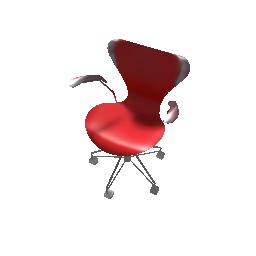} &
	\includegraphics[width=\gfxwidthB\linewidth,trim={1.cm 1.cm 1.cm 1.cm},clip]{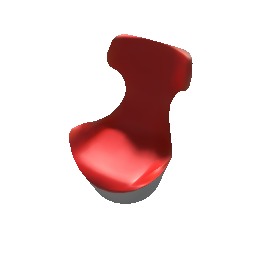}\\
	Input Image & \multicolumn{4}{c}{Surface Light Field Transfer} \\
\end{tabular}
    \caption{\textbf{Generative Model.}
    We show random samples from our generative model (first row), a latent space interpolation (second row), and surface light field transfers to different geometries.
    }
	\label{fig:gen_rand}
\end{figure}

In addition to predicting appearance from a single image, we use our approach as a generative model for learning the conditional distribution of surface light fields given a 3D model.
In \figref{fig:gen_rand}, we show the output of our model for randomly sampled appearance codes~$\bz$ as well as latent space interpolations using our VAE-based generative model.
Our model learns a meaningful latent space, which also allows for transferring appearance from one model to another.
We observe that the predictions contain plausible specular reflections and shadows. %

\section{Conclusion}
\label{Sec:Conclu}

We proposed conditional implicit surface light fields, a novel representation of visual appearance of 3D objects.
Our model reasons about complex visual effects including texture, diffuse and specular reflection as well as shadows.
Our representation allows for inferring arbitrary viewpoints with novel light configurations.
We find that conditional implicit surface light fields are an effective representation of visual appearance.
In future work, we plan to further increase the representation power of this model by exploiting convolutional features.

\vspace{-0.3cm}
\subsection*{Acknowledgements}
This work was supported by an NVIDIA research gift. The authors thank the International Max Planck Research School for Intelligent Systems (IMPRS-IS) for supporting Michael Niemeyer.

\clearpage
\bibliographystyle{splncs04}
\bibliography{bibliography_long,bibliography,bibliography_custom}
\newpage
\setcounter{section}{0}
\section*{Supplementary Material}
\begin{abstract} 
    In this supplementary material, we first provide implementation details regarding our network architectures and training procedure in~\secref{sec:implementation-details}.
Next, we explain our rendering setup in greater detail in~\secref{sec:rendering-setup}.
In~\secref{sec:environmentmaps}, we discuss how our method can be used in combination with environment maps. 
Additional experimental results can be found in~\secref{sec:more-results}.

\end{abstract}

\section{Implementation Details}
\label{sec:implementation-details}

We first describe our network architectures in~\secref{sec:supmat} and provide details regarding the training procedure in~\secref{subsec:training-details}.

\subsection{Network Architectures}
\label{sec:supmat}
\begin{figure} 
    \centering
    \includegraphics[width=0.48\linewidth]{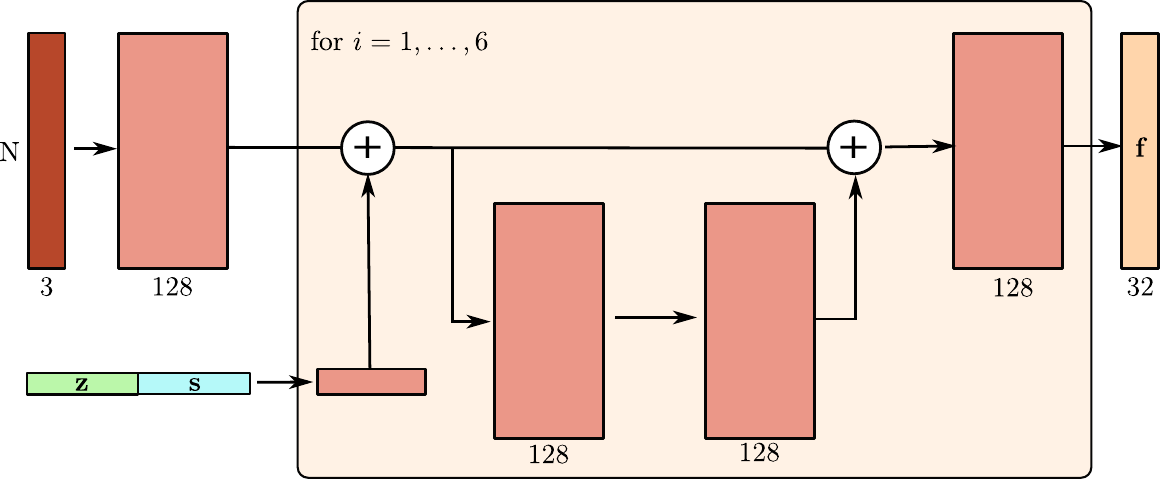}
    \quad
    \includegraphics[width=0.48\linewidth]{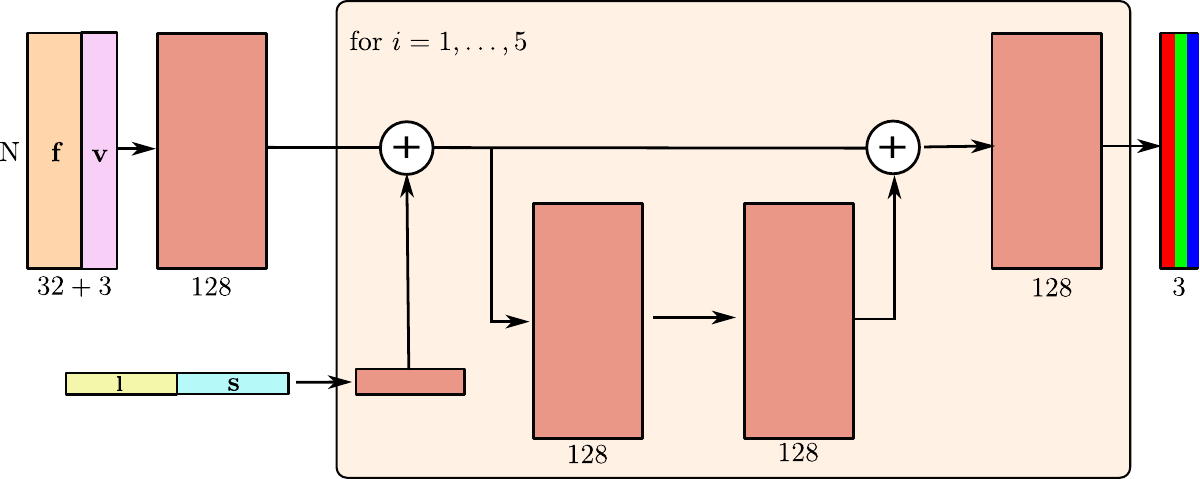}
    \caption{\textbf{2-Step Model.} Network architectures of our 2-step model that consists of two separate networks. We pass the 3D locations through the first network to determine the appearance feature vector $\mathbf{f}$. For the single view experiment,
    we condition this network on the input image $(\bz)$ and the object shape $(\bs)$. The second network maps $\mathbf{f}$, the view direction~$\bv$, the light configuration~$\bl$ and the shape encoding~$\bs$ to an RGB value. 
    }
    \label{fig:2-step}
\end{figure}

\boldparagraph{2-Step Model}
In \figref{fig:2-step}, we show the network architectures of our 2-Step model. 
The first network predicts a 32-dimensional appearance feature vector $\mathbf{f}$ for a given 3D location~$\bp$.
In the single view experiment, we condition the network on the geometry encoding~$\bs$ and input image encoding~$\bz$ as discussed in the main paper.
The second network maps the appearance feature vector $\mathbf{f}$, the view direction $\bv$, the light configuration~$\bl$ (\eg the light location), and the 3D geometry encoding~$\bs$ to an RGB value.

The networks consist of 6 (first network) and 5 (second network) fully-connected residual blocks with a hidden dimension of 128.
For the single view experiment, we concatenate all input encodings to form a single vector which we pass through a fully-connected layer and add to the output of every block.
 
\begin{figure}[h]
    \centering
    \includegraphics[width=0.7\linewidth]{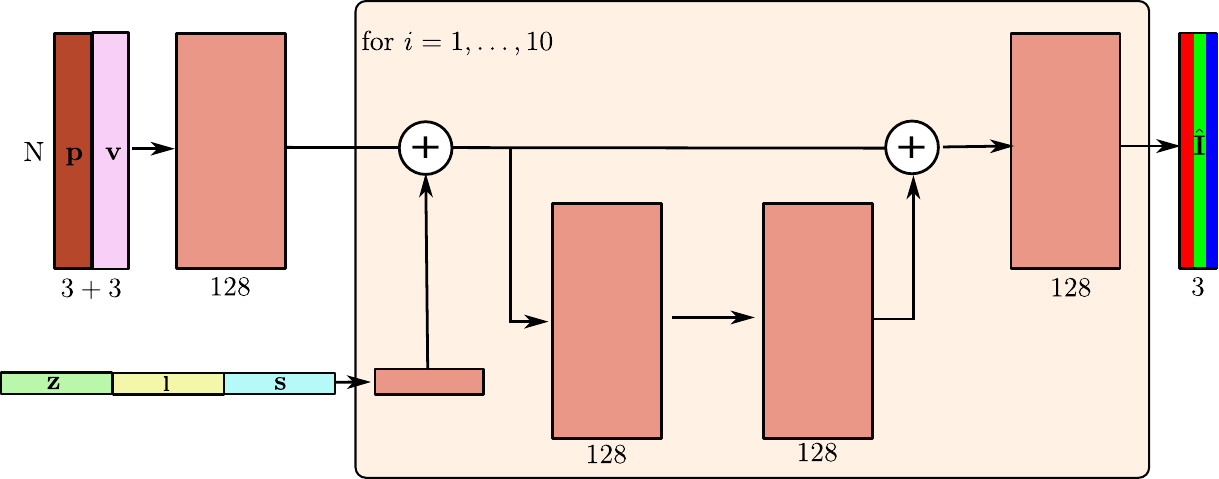}
    \caption{\textbf{1-Step Model.} Network architecture of the 1-step model. The network maps a 3D location $\bp$ and a view direction $\bv$ to an RGB value. We condition the network on the image encoding $\bz$, the light configuration $\bl$ and the shape encoding $\bs$.}
    \label{fig:1-step}
\end{figure}

\boldparagraph{1-Step Model}
In contrast to the 2-step model, the 1-Step model consists of a single neural network that maps a 3D location and the view direction~$\bv$ to a RGB value. 
We condition the network on the geometry encoding~$\bs$, the input image encoding~$\bz$, and the light configuration $\bl$.
Similar to before, we use 10 fully-connected residual blocks. The network architecture is illustrated in~\figref{fig:1-step}.

\begin{figure}
    \centering
    \includegraphics[width=0.95\linewidth]{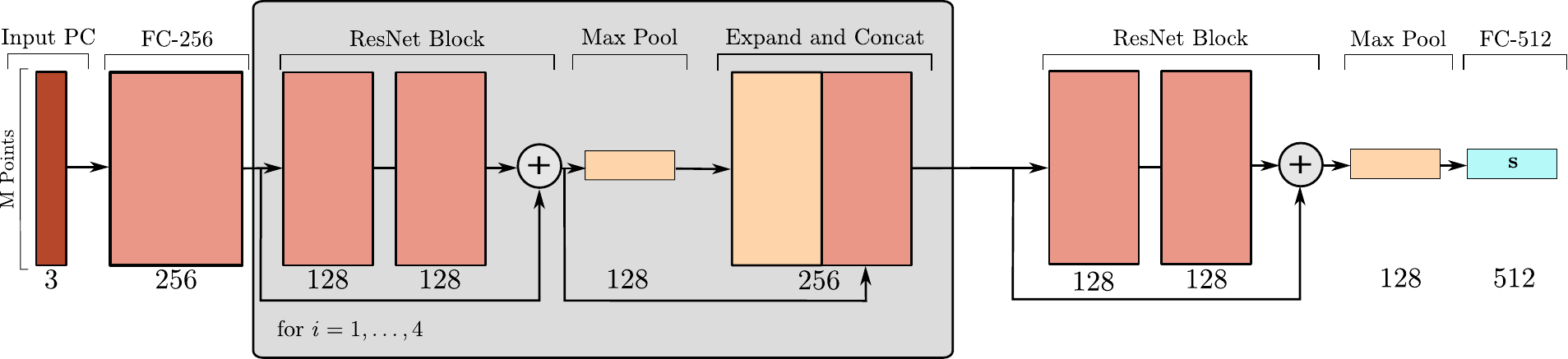}
    \caption{\textbf{Geometry Encoder.} We show the architecture of the geometry encoder. The geometry encoder maps a randomly sampled pointcloud of the input object to a global feature vector $\bs$.}
    \label{fig:geo_encoder}
\end{figure}

\boldparagraph{Geometry Encoder}
For encoding the geometry of the input object, we randomly sample 2048 points on the surface of the object.
We use a PointNet architecture~\cite{Qi2017CVPR} with residual blocks and intermediate pooling \cite{Mescheder2019CVPR} to obtain a global feature vector $\bs$. 
The network architecture is shown in \figref{fig:geo_encoder}. 

\begin{figure}
    \centering
    \includegraphics[width=0.95\linewidth]{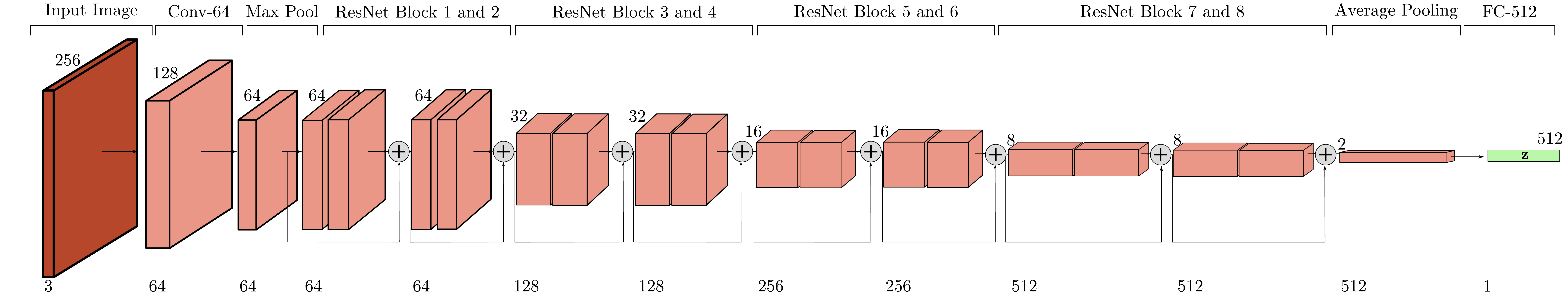}
    \caption{\textbf{Image Encoder.} We use a pre-trained Resnet18 for encoding the input image into a global feature vector $\bz$.}
    \label{fig:image_encoder}
\end{figure}

\boldparagraph{Image Encoder}
We used a Resnet18~\cite{He2016CVPR} network pre-trained on ImageNet~\cite{Krizhevsky2012NIPS} for encoding the input image into a global feature encoding $\bz$. 
Our input images have a resolution of $256^2$ pixels and the image encoding dimension is $512$.
A detailed visualization is provided in \figref{fig:image_encoder}.

\begin{figure}
    \centering
    \includegraphics[width=0.95\linewidth]{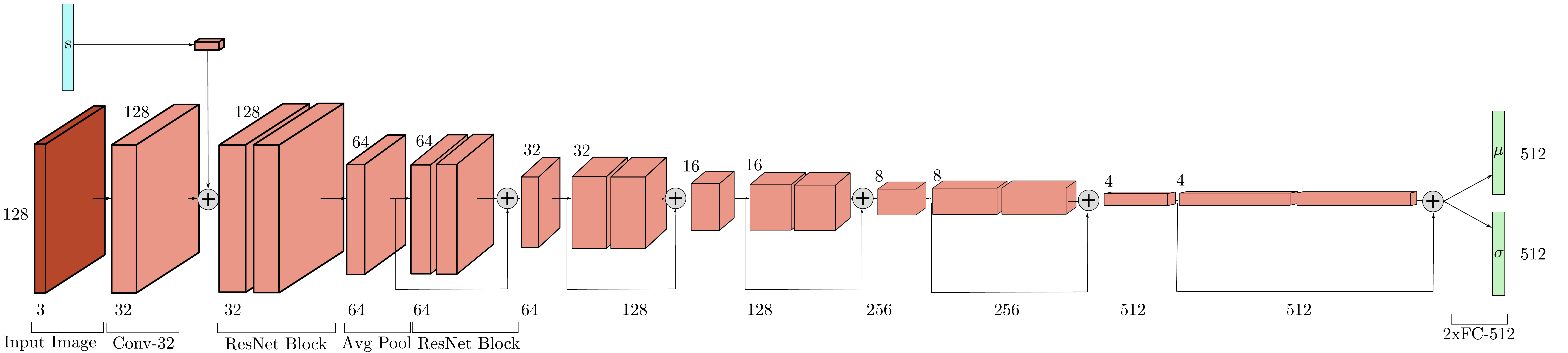}
    \caption{\textbf{VAE Encoder.} Our VAE encoder maps an image and a geometry encoding to the mean and variance of the Gaussian distributed latent code~$\bz$.}
    \label{fig:vae_encoder}
\end{figure}

\boldparagraph{VAE Encoder}
We implement our generative model as a variational autoencoder \cite{Kingma2014ICLR}.
As encoder, we use a Resnet18-based architecture \cite{He2016CVPR} conditioned on the shape encoding~$\bs$.
The network maps an image~$\bI$ as well as the corresponding shape encoding~$\bs$ to a mean and a variance vector of a Gaussian distribution (see \figref{fig:vae_encoder}). 
We sample the latent code $\bz$ from this Gaussian distribution.

\begin{figure}
    \centering
    \includegraphics[width=0.95\linewidth]{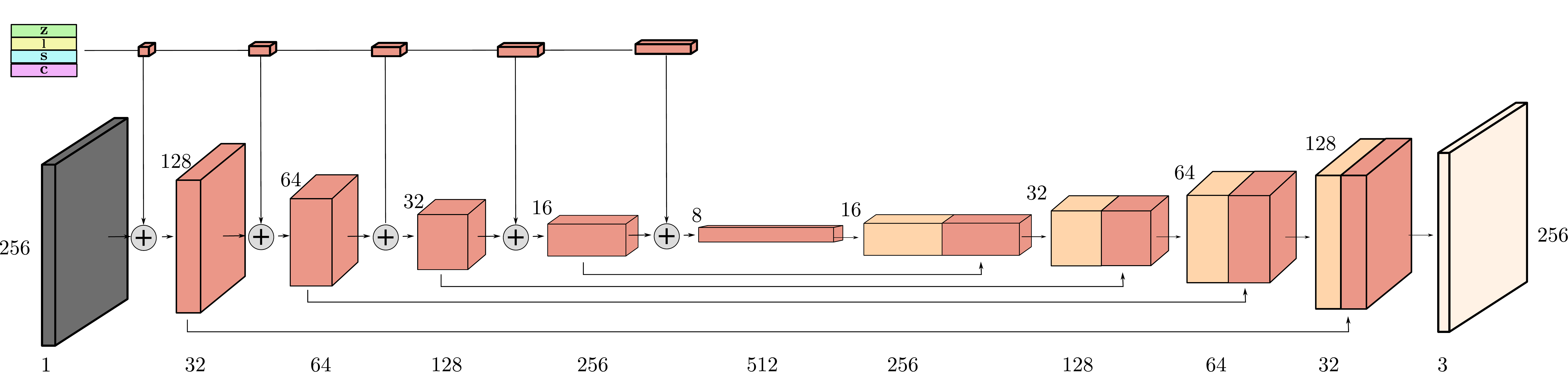}
    \caption{\textbf{Image-to-Image Translation.} We show the architecture of the Image-to-Image Translation baseline. We pass a depth image through a U-net architecture to directly infer an RGB image.
    }
    \label{fig:unet}
\end{figure}

\boldparagraph{Image-to-Image Translation Baseline}
In the single view experiment, we use an image-to-image translation approach with a U-Net architecture~\cite{Ronneberger2015MICCAI} as baseline.
The network maps a depth map to an RGB image and is conditioned on the geometry encoding $\bs$,  light configuration $\bl$, camera position $\bc$ and the image encoding $\bz$ (see \figref{fig:unet}).
The network is trained in a supervised fashion with paired examples of depth maps and RGB images.
During training, we use the same loss function as for our model which is discussed in the main paper.

\subsection{Training Details}
\label{subsec:training-details}
For simplicity, we explain the training procedure of a single training iteration considering one element of the batch.
Depending on the experiment, we first encode the provided conditions into encodings $\bs$, $\bz$, or $\bl$.
Second, we sample a set of 500 (single view experiment) and 2048 (single object experiment on chair) pixels of the GT depth map that are inside the object region and map these pixels to their 3D locations using camera intrinsics.
We then set the view directions $\bv$ to the normalized vector from the 3D locations to the camera center.
Next, we use the 3D locations, view directions and the encodings to obtain the corresponding RGB values.
We then determine the loss between the predicted and the ground truth RGB values. %
In all experiments, we use the ADAM optimizer~\cite{Kingma2015ICLR} with a learning rate of $10^{-4}$.
For the single view experiment and the generative model, we train our models with a batch size of 16.

\section{Rendering Setup}
\label{sec:rendering-setup}
In this section, we describe the rendering setups in greater detail.
We used the Blender Cycles render engine~\cite{Blender} for rendering  3D objects from the Photoshape dataset~\cite{Park2018SIG} as discussed in the main paper. %
The 3D objects are in canonical pose and scaled to the unit cube.
We randomly sample camera locations on the northern hemisphere with radius 1.5 (blender units).
Similarly, we sample light locations on the northern hemisphere with radius 10. %
The number of light locations per view differs across the experiments and is hence given in the respective section.
For the single view experiment, we additionally use a uniform background light of strength $0.2$ to guarantee that the input images are not completely black.  %
For the reflection analysis experiment, we sample the color of the light source uniformly in RGB space.
For all other experiments, we use a white light source.

\section{Spatially Varying Illumination}
\label{sec:environmentmaps}
To determine the surface light field \wrt environment maps, we approximate the illumination with a set of small light sources. 
We determine equidistributed points on a sphere using the method from~\cite{Deserno2004}. 
We query the color at each of these points from the corresponding pixels in the environment map and consider each point as a single point light source.
For each light source, we predict a separate RGB image and combine the resulting images as described in the main paper.

As our rendering procedure of the GT images contains a transformation from a radiometrically linear color space to the sRGB space\footnote{\url{https://docs.blender.org/manual/en/latest/render/color_management.html}}, 
the predictions of our model are also in the non-linear sRGB space.
For composing predicted images, it is thus necessary to transform them into a linear color space \cite{Anderson1996MCS} before adding the RGB values.
In the linear space, we add the images channel- and pixel-wise and then multiply the sum with a constant value reducing overexposure. 
The constant value is chosen to have a mean intensity of 0.35 in the image.
Intuitively, this corresponds to varying the shutter time of a real camera.
Finally, we transform the result back to the sRGB space.

\section{Additional Results}
\label{sec:more-results}
\begin{figure}[h]
    \centering
    \begin{tabular}{cc@{}c@{}c@{}c}
	\includegraphics[width=0.19\linewidth,trim={.75cm .5cm .75cm .5cm},clip]{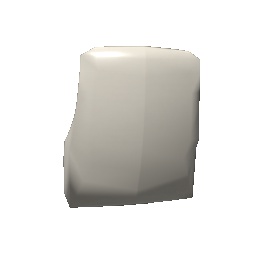} &
	\includegraphics[width=0.19\linewidth,trim={.75cm .5cm .75cm .cm},clip]{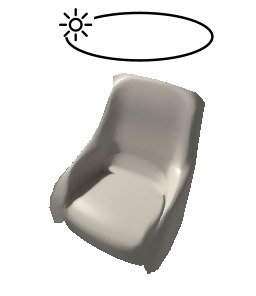} &
	\includegraphics[width=0.19\linewidth,trim={.75cm .5cm .75cm .cm},clip]{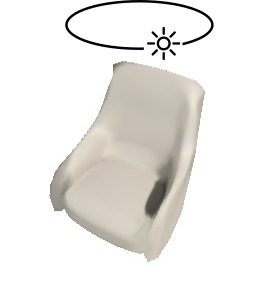} &
	\includegraphics[width=0.19\linewidth,trim={.75cm .5cm .75cm .cm},clip]{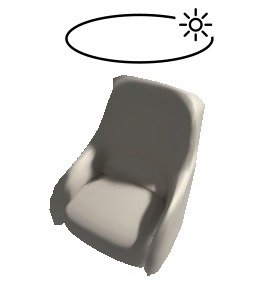} &
	\includegraphics[width=0.19\linewidth,trim={.75cm .5cm .75cm .cm},clip]{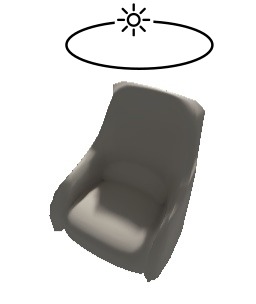} \\
	\includegraphics[width=0.19\linewidth,trim={.75cm .5cm .75cm .5cm},clip]{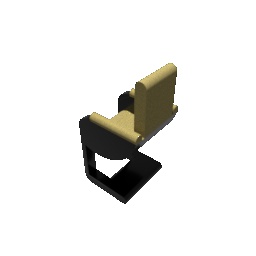} &
	\includegraphics[width=0.19\linewidth,trim={.75cm .5cm .75cm .5cm},clip]{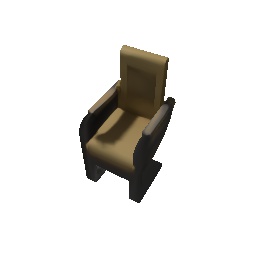} &
	\includegraphics[width=0.19\linewidth,trim={.75cm .5cm .75cm .5cm},clip]{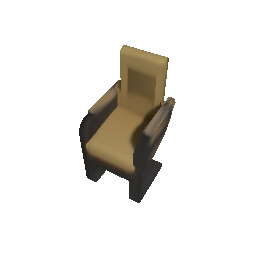} &
	\includegraphics[width=0.19\linewidth,trim={.75cm .5cm .75cm .5cm},clip]{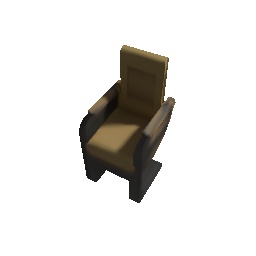} &
	\includegraphics[width=0.19\linewidth,trim={.75cm .5cm .75cm .5cm},clip]{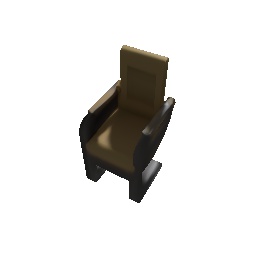} \\
	\includegraphics[width=0.19\linewidth,trim={.75cm .5cm .75cm .5cm},clip]{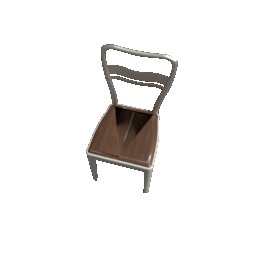} &
	\includegraphics[width=0.19\linewidth,trim={.75cm .5cm .75cm .5cm},clip]{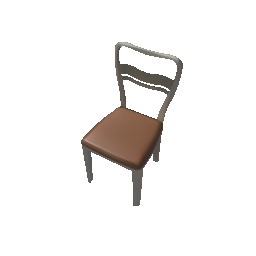} &
	\includegraphics[width=0.19\linewidth,trim={.75cm .5cm .75cm .5cm},clip]{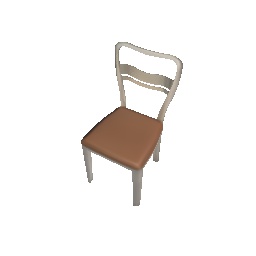} &
	\includegraphics[width=0.19\linewidth,trim={.75cm .5cm .75cm .5cm},clip]{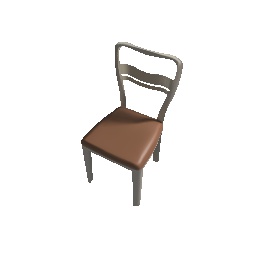} &
	\includegraphics[width=0.19\linewidth,trim={.75cm .5cm .75cm .5cm},clip]{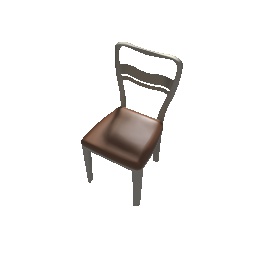} \\
	\includegraphics[width=0.19\linewidth,trim={.75cm .5cm .75cm .5cm},clip]{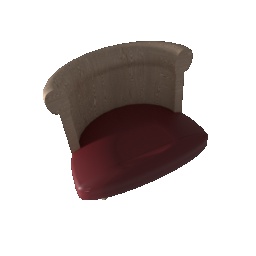} &
	\includegraphics[width=0.19\linewidth,trim={.75cm .5cm .75cm .5cm},clip]{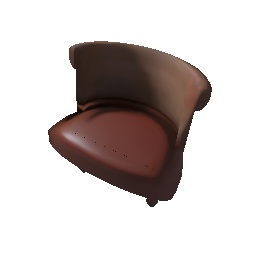} &
	\includegraphics[width=0.19\linewidth,trim={.75cm .5cm .75cm .5cm},clip]{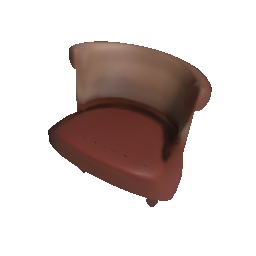} &
	\includegraphics[width=0.19\linewidth,trim={.75cm .5cm .75cm .5cm},clip]{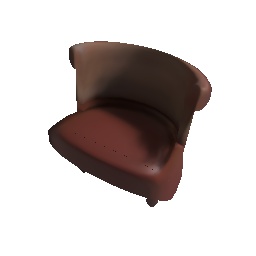} &
	\includegraphics[width=0.19\linewidth,trim={.75cm .5cm .75cm .5cm},clip]{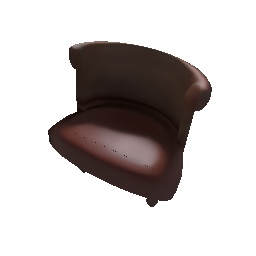} \\
	\includegraphics[width=0.19\linewidth,trim={.75cm .5cm .75cm .5cm},clip]{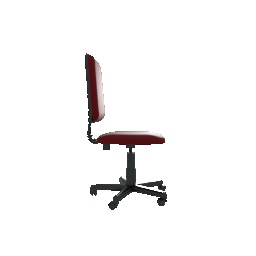} &
	\includegraphics[width=0.19\linewidth,trim={.75cm .5cm .75cm .5cm},clip]{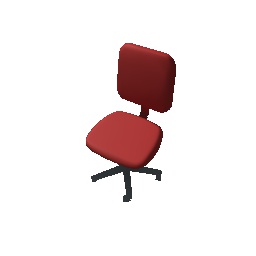} &
	\includegraphics[width=0.19\linewidth,trim={.75cm .5cm .75cm .5cm},clip]{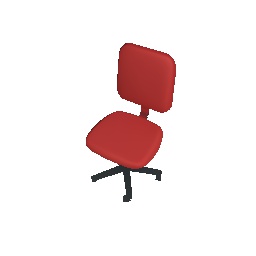} &
	\includegraphics[width=0.19\linewidth,trim={.75cm .5cm .75cm .5cm},clip]{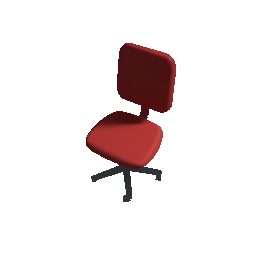} &
	\includegraphics[width=0.19\linewidth,trim={.75cm .5cm .75cm .5cm},clip]{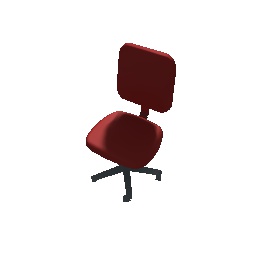} \\
	\includegraphics[width=0.19\linewidth,trim={.75cm .5cm .75cm .5cm},clip]{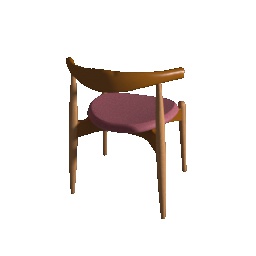} &
	\includegraphics[width=0.19\linewidth,trim={.75cm .5cm .75cm .5cm},clip]{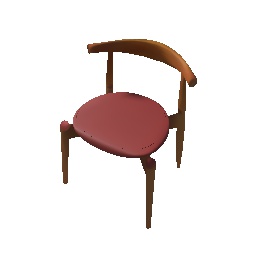} &
	\includegraphics[width=0.19\linewidth,trim={.75cm .5cm .75cm .5cm},clip]{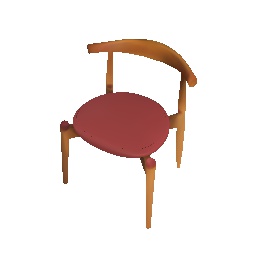} &
	\includegraphics[width=0.19\linewidth,trim={.75cm .5cm .75cm .5cm},clip]{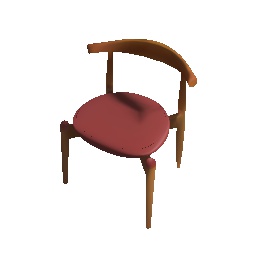} &
	\includegraphics[width=0.19\linewidth,trim={.75cm .5cm .75cm .5cm},clip]{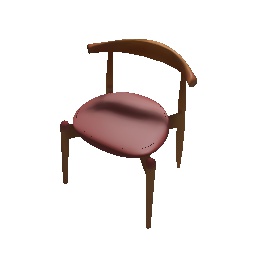} \\
	Input Image & \multicolumn{4}{c}{Predictions}
\end{tabular} 
    \caption{\textbf{Single View Appearance Prediction.} Results of our 2-step model for predicting the appearance from a single image of an unseen object. We show predictions for four different light configurations.
    }
	\label{fig:sv_gt} 
\end{figure}

\begin{figure}[h]   
    \centering
    \begin{tabular}{cc@{}c@{}c@{}c}
	\includegraphics[width=0.19\linewidth,trim={.75cm .5cm .75cm .5cm},clip]{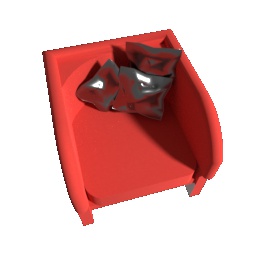} &
	\includegraphics[width=0.19\linewidth,trim={0cm 0cm 0cm 0cm},clip]{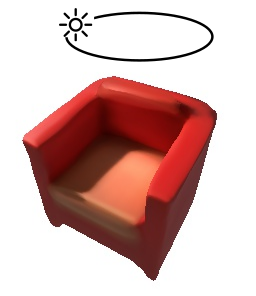} &
	\includegraphics[width=0.19\linewidth,trim={0cm, 0cm 0cm 0cm},clip]{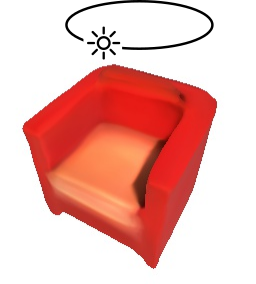} &
	\includegraphics[width=0.19\linewidth,trim={0cm 0cm 0cm 0cm},clip]{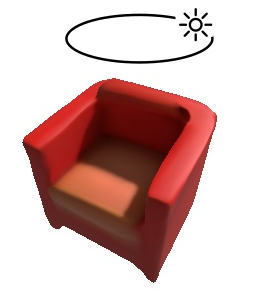} &
	\includegraphics[width=0.19\linewidth,trim={.0cm 0cm 0cm 0cm},clip]{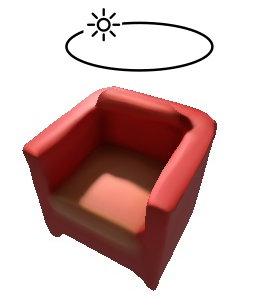} \\
	\includegraphics[width=0.19\linewidth,trim={.75cm .5cm .75cm .cm},clip]{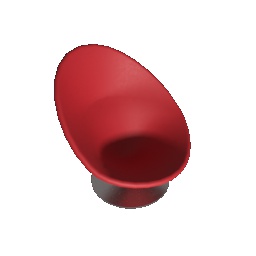} &
	\includegraphics[width=0.19\linewidth,trim={.75cm .5cm .75cm .5cm},clip]{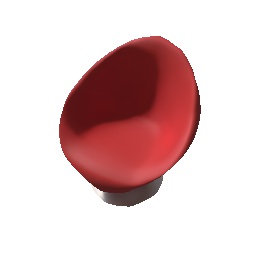} &
	\includegraphics[width=0.19\linewidth,trim={.75cm .5cm .75cm .5cm},clip]{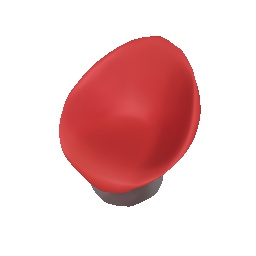} &
	\includegraphics[width=0.19\linewidth,trim={.75cm .5cm .75cm .5cm},clip]{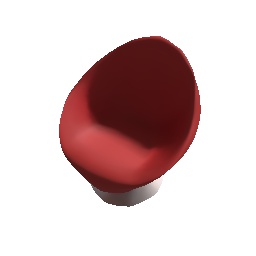} &
	\includegraphics[width=0.19\linewidth,trim={.75cm .5cm .75cm .5cm},clip]{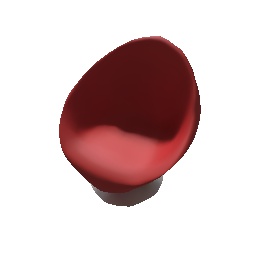} \\
	\includegraphics[width=0.19\linewidth,trim={.75cm .5cm .75cm .5cm},clip]{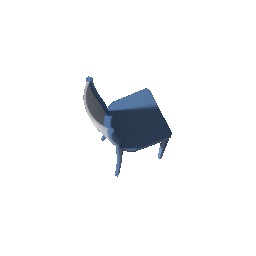} &
	\includegraphics[width=0.19\linewidth,trim={.75cm .5cm .75cm .5cm},clip]{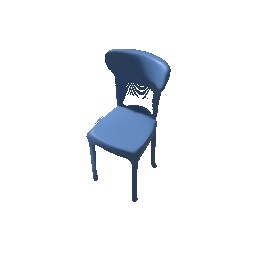} &
	\includegraphics[width=0.19\linewidth,trim={.75cm .5cm .75cm .5cm},clip]{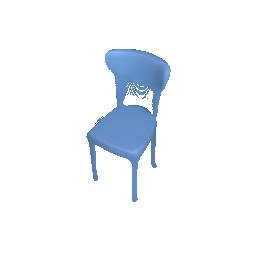} &
	\includegraphics[width=0.19\linewidth,trim={.75cm .5cm .75cm .5cm},clip]{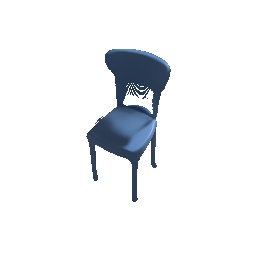} &
	\includegraphics[width=0.19\linewidth,trim={.75cm .5cm .75cm .5cm},clip]{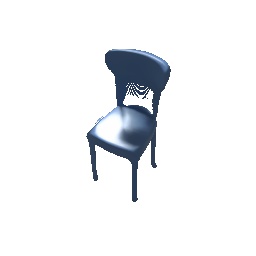} \\
	\includegraphics[width=0.19\linewidth,trim={.75cm .5cm .75cm .5cm},clip]{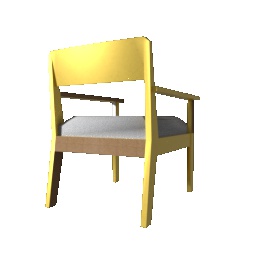} &
	\includegraphics[width=0.19\linewidth,trim={.75cm .5cm .75cm .5cm},clip]{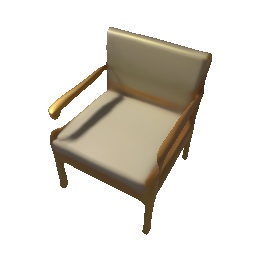} &
	\includegraphics[width=0.19\linewidth,trim={.75cm .5cm .75cm .5cm},clip]{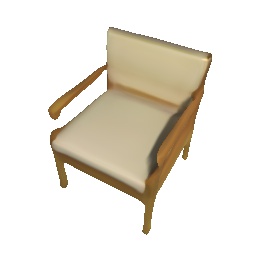} &
	\includegraphics[width=0.19\linewidth,trim={.75cm .5cm .75cm .5cm},clip]{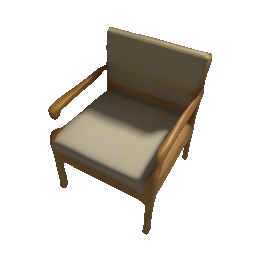} &
	\includegraphics[width=0.19\linewidth,trim={.75cm .5cm .75cm .5cm},clip]{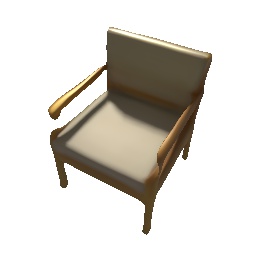} \\
	\includegraphics[width=0.19\linewidth,trim={.75cm .5cm .75cm .5cm},clip]{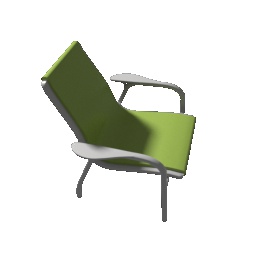} &
	\includegraphics[width=0.19\linewidth,trim={.75cm .5cm .75cm .5cm},clip]{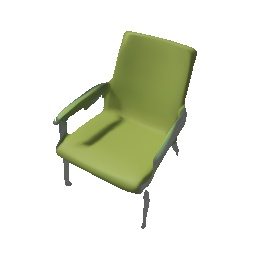} &
	\includegraphics[width=0.19\linewidth,trim={.75cm .5cm .75cm .5cm},clip]{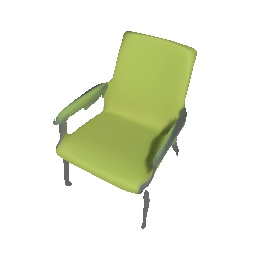} &
	\includegraphics[width=0.19\linewidth,trim={.75cm .5cm .75cm .5cm},clip]{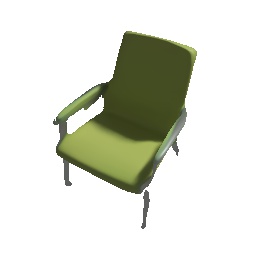} &
	\includegraphics[width=0.19\linewidth,trim={.75cm .5cm .75cm .5cm},clip]{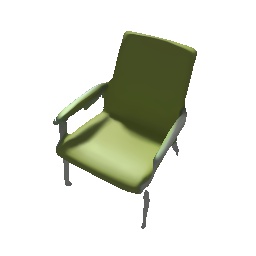} \\
	\includegraphics[width=0.19\linewidth,trim={.75cm .5cm .75cm .5cm},clip]{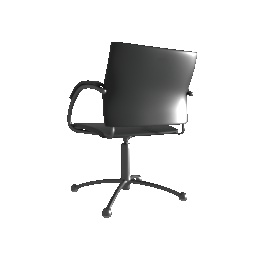} &
	\includegraphics[width=0.19\linewidth,trim={.75cm .5cm .75cm .5cm},clip]{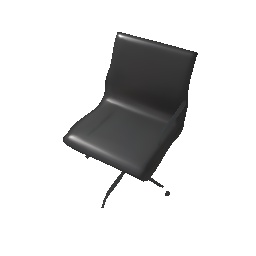} &
	\includegraphics[width=0.19\linewidth,trim={.75cm .5cm .75cm .5cm},clip]{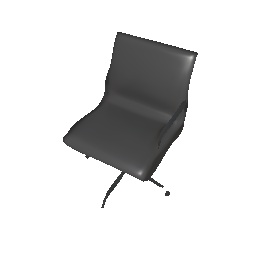} &
	\includegraphics[width=0.19\linewidth,trim={.75cm .5cm .75cm .5cm},clip]{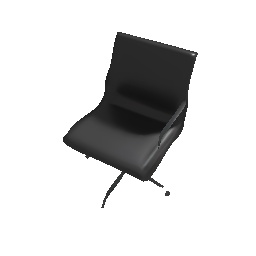} &
	\includegraphics[width=0.19\linewidth,trim={.75cm .5cm .75cm .5cm},clip]{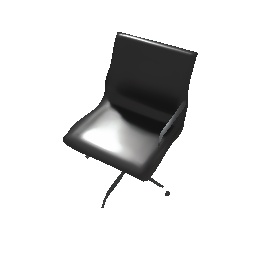} \\
	Input Image & \multicolumn{4}{c}{Predictions}
\end{tabular} 
    \caption{\textbf{Image-based Reconstruction.} We show reconstructions of both geometry and appearance from a single input image for four different light configurations.}
	\label{fig:sv_dvr} 
\end{figure}

\begin{figure}[h]
    \centering
    \begin{tabular}{c@{}c@{}c@{}c@{}c}
	\includegraphics[width=0.142\linewidth,trim={0cm 0cm 0cm 0cm},clip]{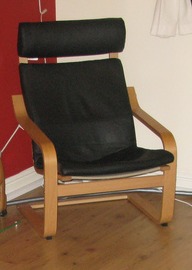} &
	\includegraphics[width=0.21\linewidth,trim={0.5cm 1cm 0.5cm 0cm},clip]{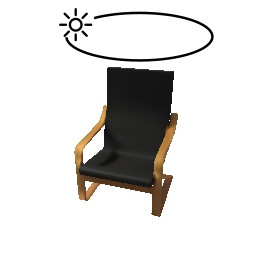} &
	\includegraphics[width=0.21\linewidth,trim={0.5cm 1cm 0.5cm 0cm},clip]{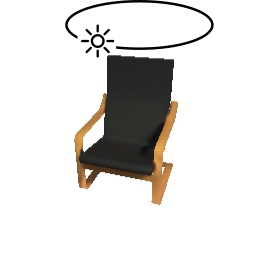} &
	\includegraphics[width=0.21\linewidth,trim={0.5cm 1cm 0.5cm 0cm},clip]{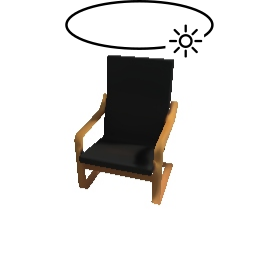} &
	\includegraphics[width=0.21\linewidth,trim={0.5cm 1cm 0.5cm 0cm},clip]{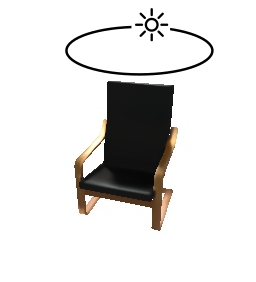} \\
	\includegraphics[width=0.142\linewidth,trim={0.1cm .1cm .1cm .1cm},clip]{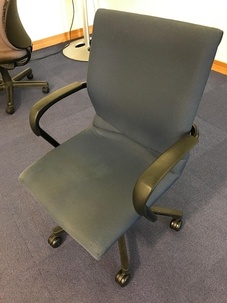} &
	\includegraphics[width=0.21\linewidth,trim={1.5cm 1.5cm 1.5cm 1.5cm},clip]{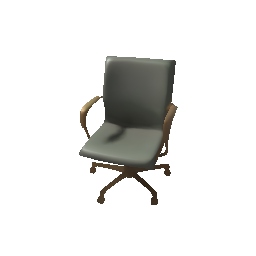} &
	\includegraphics[width=0.21\linewidth,trim={1.5cm 1.5cm 1.5cm 1.5cm},clip]{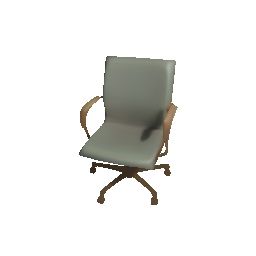} &
	\includegraphics[width=0.21\linewidth,trim={1.5cm 1.5cm 1.5cm 1.5cm},clip]{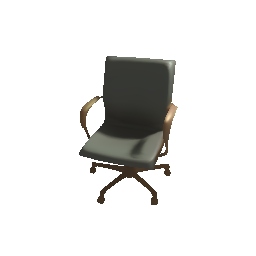} &
	\includegraphics[width=0.21\linewidth,trim={1.5cm 1.5cm 1.5cm 1.5cm},clip]{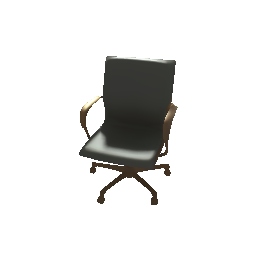} \\
	\includegraphics[width=0.142\linewidth,trim={0.1cm .1cm .1cm .1cm},clip]{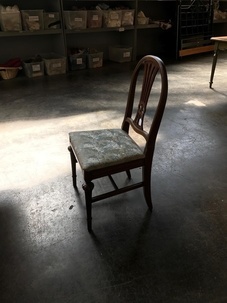} &
	\includegraphics[width=0.21\linewidth,trim={1.5cm 1.5cm 1.5cm 1.5cm},clip]{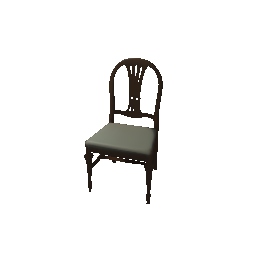} &
	\includegraphics[width=0.21\linewidth,trim={1.5cm 1.5cm 1.5cm 1.5cm},clip]{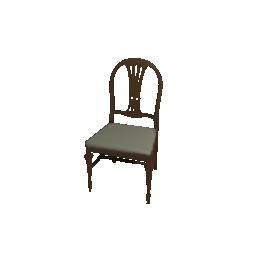} &
	\includegraphics[width=0.21\linewidth,trim={1.5cm 1.5cm 1.5cm 1.5cm},clip]{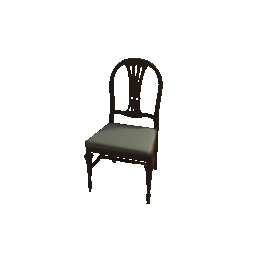} &
	\includegraphics[width=0.21\linewidth,trim={1.5cm 1.5cm 1.5cm 1.5cm},clip]{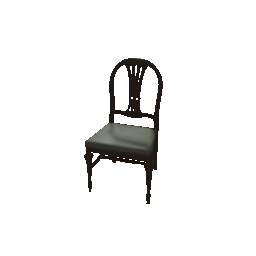} \\
	\includegraphics[width=0.142\linewidth,trim={0.1cm .1cm .1cm .1cm},clip]{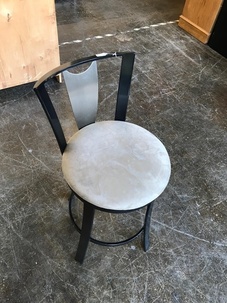} &
	\includegraphics[width=0.21\linewidth,trim={1.5cm 1.5cm 1.5cm 1.5cm},clip]{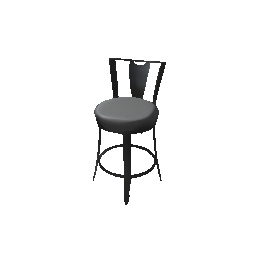} &
	\includegraphics[width=0.21\linewidth,trim={1.5cm 1.5cm 1.5cm 1.5cm},clip]{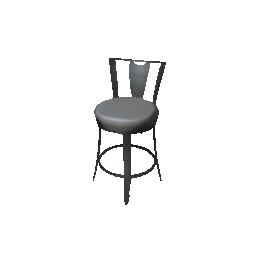} &
	\includegraphics[width=0.21\linewidth,trim={1.5cm 1.5cm 1.5cm 1.5cm},clip]{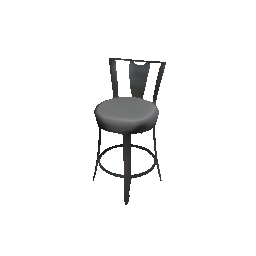} &
	\includegraphics[width=0.21\linewidth,trim={1.5cm 1.5cm 1.5cm 1.5cm},clip]{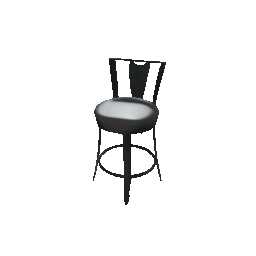} \\
	\includegraphics[width=0.142\linewidth,trim={0.1cm .1cm .1cm .1cm},clip]{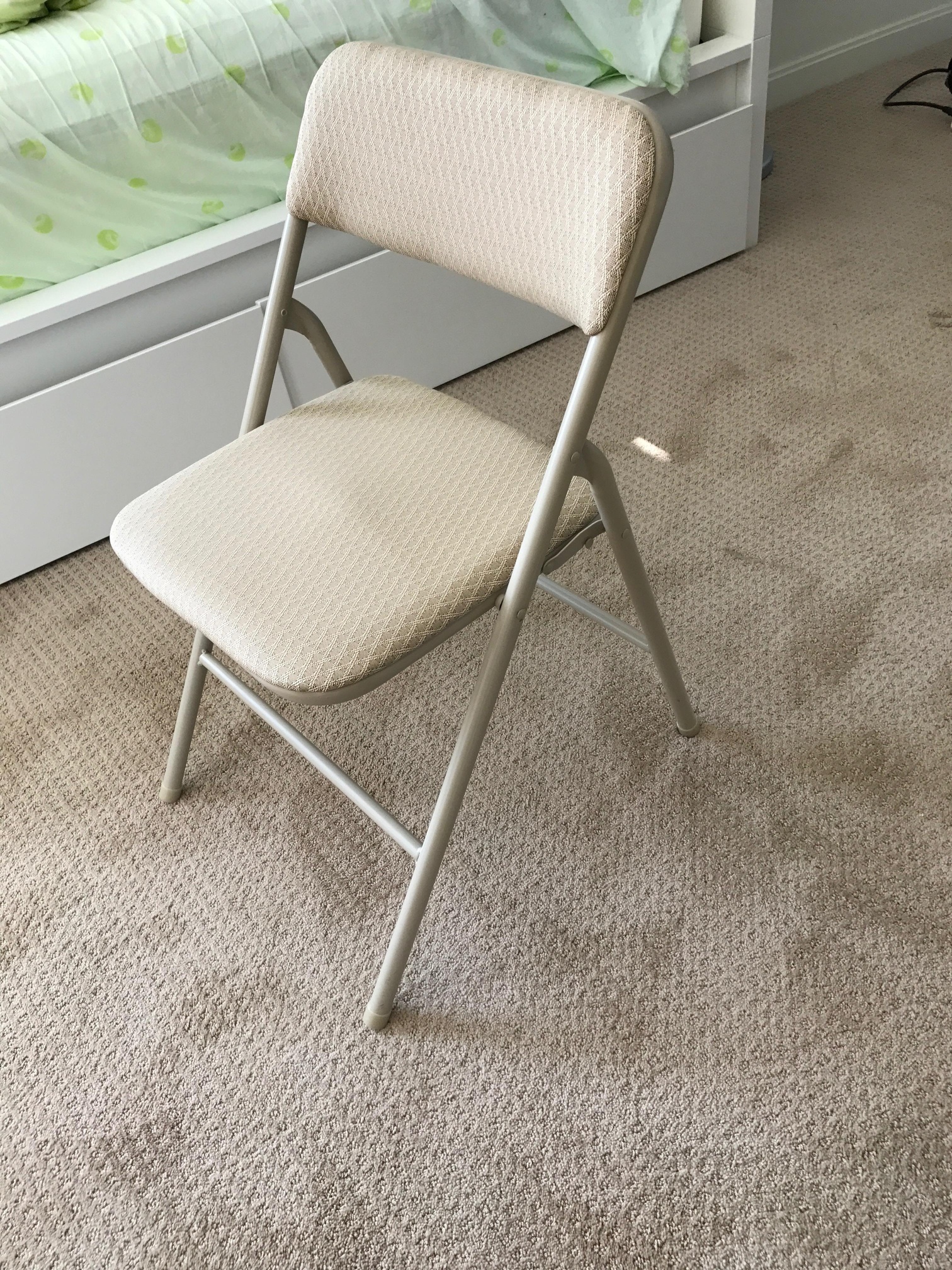} &
	\includegraphics[width=0.21\linewidth,trim={1.5cm 1.5cm 1.5cm 1.5cm},clip]{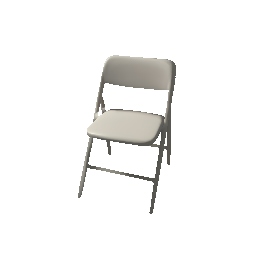} &
	\includegraphics[width=0.21\linewidth,trim={1.5cm 1.5cm 1.5cm 1.5cm},clip]{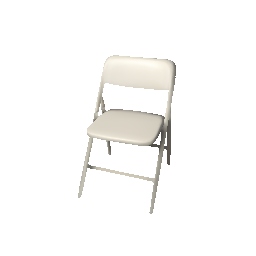} &
	\includegraphics[width=0.21\linewidth,trim={1.5cm 1.5cm 1.5cm 1.5cm},clip]{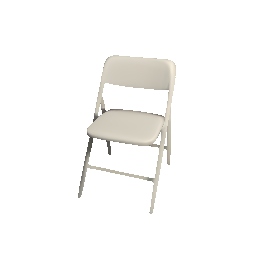} &
	\includegraphics[width=0.21\linewidth,trim={1.5cm 1.5cm 1.5cm 1.5cm},clip]{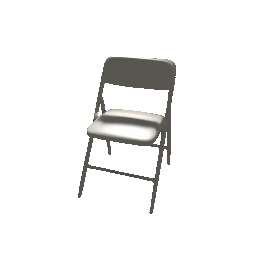} \\
\end{tabular}
 
    \caption{\textbf{Real Images.} Results of our method when using real images of unseen objects as input. We use images and corresponding 3D geometries from the Pix3D dataset \cite{SunPix3D2018CVPR}.}
	\label{fig:sv_pix3d} 
\end{figure}

\begin{figure}[h]
    \centering
    \def\gfxwidthA{0.166}
\begin{tabular}{c@{}c@{}c@{}c@{}c@{}c@{}c@{}c}
	\includegraphics[width=\gfxwidthA\linewidth,trim={1.cm 1.cm 1.cm 1.cm},clip]{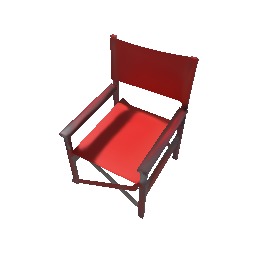} &
	\includegraphics[width=\gfxwidthA\linewidth,trim={1.cm 1.cm 1.cm 1.cm},clip]{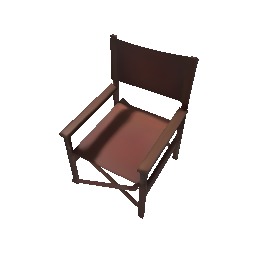} & 
	\includegraphics[width=\gfxwidthA\linewidth,trim={1.cm 1.cm 1.cm 1.cm},clip]{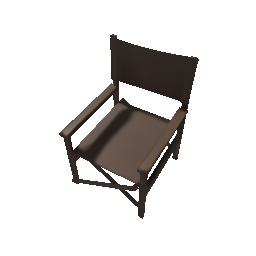} &
	\includegraphics[width=\gfxwidthA\linewidth,trim={1.cm 1.cm 1.cm 1.cm},clip]{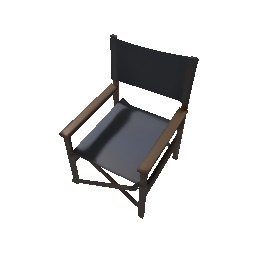} &
	\includegraphics[width=\gfxwidthA\linewidth,trim={1.cm 1.cm 1.cm 1.cm},clip]{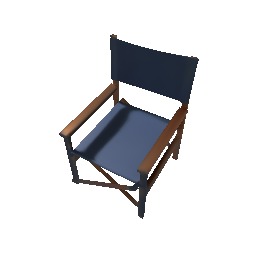}&
	\includegraphics[width=\gfxwidthA\linewidth,trim={1.cm 1.cm 1.cm 1.cm},clip]{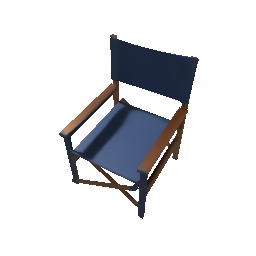}\\
	\includegraphics[width=\gfxwidthA\linewidth,trim={1.cm 1.cm 1.cm 1.cm},clip]{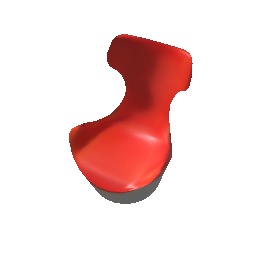} &
	\includegraphics[width=\gfxwidthA\linewidth,trim={1.cm 1.cm 1.cm 1.cm},clip]{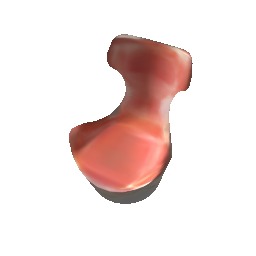} & 
	\includegraphics[width=\gfxwidthA\linewidth,trim={1.cm 1.cm 1.cm 1.cm},clip]{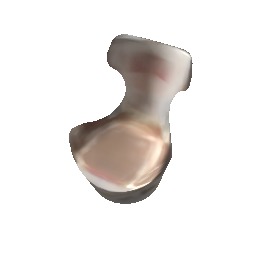} &
	\includegraphics[width=\gfxwidthA\linewidth,trim={1.cm 1.cm 1.cm 1.cm},clip]{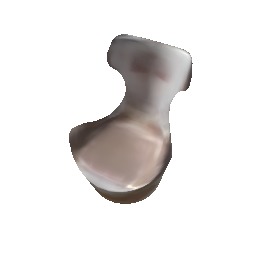} &
	\includegraphics[width=\gfxwidthA\linewidth,trim={1.cm 1.cm 1.cm 1.cm},clip]{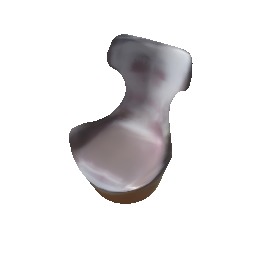}&
	\includegraphics[width=\gfxwidthA\linewidth,trim={1.cm 1.cm 1.cm 1.cm},clip]{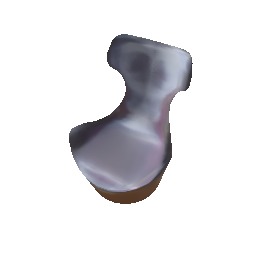}\\
	\includegraphics[width=\gfxwidthA\linewidth,trim={1.cm 1.cm 1.cm 1.cm},clip]{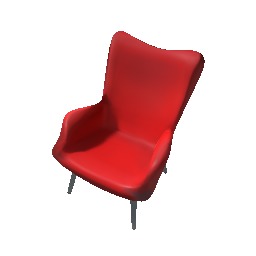} &
	\includegraphics[width=\gfxwidthA\linewidth,trim={1.cm 1.cm 1.cm 1.cm},clip]{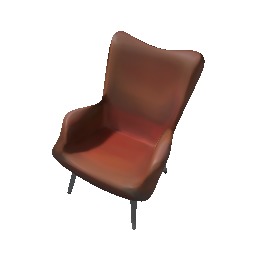} & 
	\includegraphics[width=\gfxwidthA\linewidth,trim={1.cm 1.cm 1.cm 1.cm},clip]{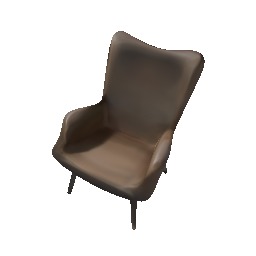} &
	\includegraphics[width=\gfxwidthA\linewidth,trim={1.cm 1.cm 1.cm 1.cm},clip]{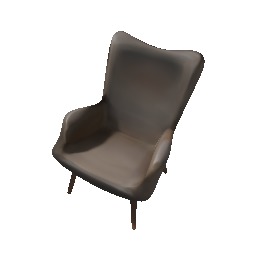} &
	\includegraphics[width=\gfxwidthA\linewidth,trim={1.cm 1.cm 1.cm 1.cm},clip]{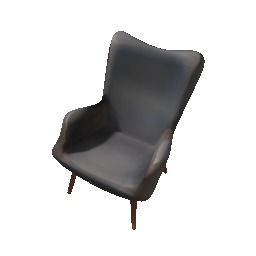}&
	\includegraphics[width=\gfxwidthA\linewidth,trim={1.cm 1.cm 1.cm 1.cm},clip]{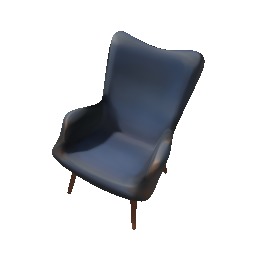}\\
\end{tabular}
 
    \caption{\textbf{Generative Model.} We show predictions of our generative model for interpolated latent codes. The light location is fixed for all predictions.}
	\label{fig:genmodel} 
\end{figure}

In this section, we provide more results for the single view appearance prediction experiments with ground truth geometry (\figref{fig:sv_gt}) and with reconstructed geometry (\figref{fig:sv_dvr}).
We also show additional results for real input images in \figref{fig:sv_pix3d}. %
Additional latent space interpolations of our generative model can be found in in~\figref{fig:genmodel}.

\begin{figure}[h]
    \centering
    \begin{tabular}{c@{}c@{}c@{}c@{}c@{}c}
	Input & 1-step 2v & 2-step 2v & 1-step & 2-step & GT\\
	\includegraphics[width=0.16\linewidth,trim={.75cm .75cm .75cm .75cm},clip]{gfx/singleview/input/shape06424_white.jpg} &
	\includegraphics[width=0.16\linewidth,trim={.75cm .75cm .75cm .75cm},clip]{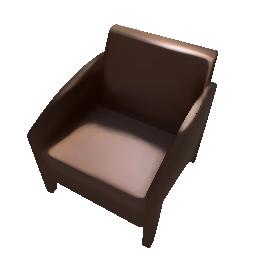} &
	\includegraphics[width=0.16\linewidth,trim={.75cm .75cm .75cm .75cm},clip]{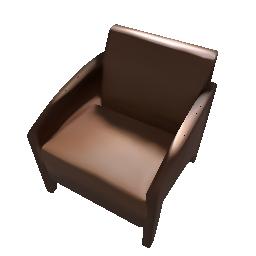} &
	\includegraphics[width=0.16\linewidth,trim={.75cm .75cm .75cm .75cm},clip]{gfx/singleview/0217_baseline_single_glob_resnet_geo_10lay_new/rot_light/shape06424/00000004_white.jpg} &
	\includegraphics[width=0.16\linewidth,trim={.75cm .75cm .75cm .75cm},clip]{gfx/singleview/0217_glob_resnet_geo/rot_light/shape06424/00000004_white.jpg} &
	\includegraphics[width=0.16\linewidth,trim={.75cm .75cm .75cm .75cm},clip]{gfx/singleview/0217_glob_resnet_geo/rot_light/shape06424/0001_0004_white.jpg} \\
	\includegraphics[width=0.16\linewidth,trim={.75cm .75cm .75cm .75cm},clip]{gfx/singleview/input/shape03629_white.jpg} &
	\includegraphics[width=0.16\linewidth,trim={.75cm .75cm .75cm .75cm},clip]{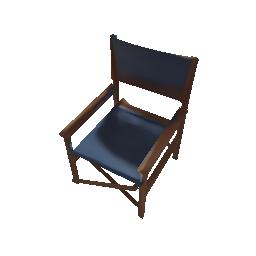} &
	\includegraphics[width=0.16\linewidth,trim={.75cm .75cm .75cm .75cm},clip]{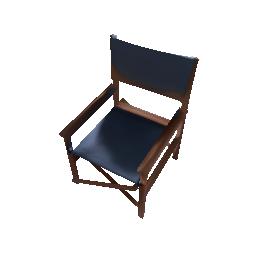} &
	\includegraphics[width=0.16\linewidth,trim={.75cm .75cm .75cm .75cm},clip]{gfx/singleview/0217_baseline_single_glob_resnet_geo_10lay_new/rot_light/shape03629/00000004_white.jpg} &
	\includegraphics[width=0.16\linewidth,trim={.75cm .75cm .75cm .75cm},clip]{gfx/singleview/0217_glob_resnet_geo/rot_light/shape03629/00000004_white.jpg} &
	\includegraphics[width=0.16\linewidth,trim={.75cm .75cm .75cm .75cm},clip]{gfx/singleview/0217_glob_resnet_geo/rot_light/shape03629/0001_0004_white.jpg} \\
	\includegraphics[width=0.16\linewidth,trim={.75cm 1.75cm .75cm 1.75cm},clip]{gfx/singleview/input/shape04410_white.jpg} &
	\includegraphics[width=0.16\linewidth,trim={.75cm 1.75cm .75cm 1.75cm},clip]{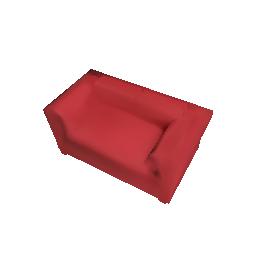} &
	\includegraphics[width=0.16\linewidth,trim={.75cm 1.75cm .75cm 1.75cm},clip]{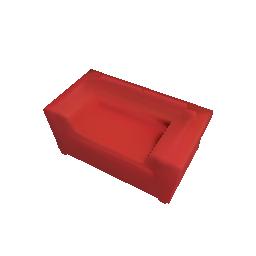} &
	\includegraphics[width=0.16\linewidth,trim={.75cm 1.75cm .75cm 1.75cm},clip]{gfx/singleview/0217_baseline_single_glob_resnet_geo_10lay_new/rot_light/shape04410/00000002_white.jpg} &
	\includegraphics[width=0.16\linewidth,trim={.75cm 1.75cm .75cm 1.75cm},clip]{gfx/singleview/0217_glob_resnet_geo/rot_light/shape04410/00000002_white.jpg} &
	\includegraphics[width=0.16\linewidth,trim={.75cm 1.75cm .75cm 1.75cm},clip]{gfx/singleview/0217_glob_resnet_geo/rot_light/shape04410/0001_0002_white.jpg} \\
	\includegraphics[width=0.16\linewidth,trim={.75cm 1.75cm .75cm .75cm},clip]{gfx/singleview/input/shape08079_white.jpg} &
	\includegraphics[width=0.16\linewidth,trim={.75cm 1.75cm .75cm .75cm},clip]{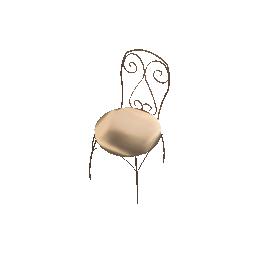} &
	\includegraphics[width=0.16\linewidth,trim={.75cm 1.75cm .75cm .75cm},clip]{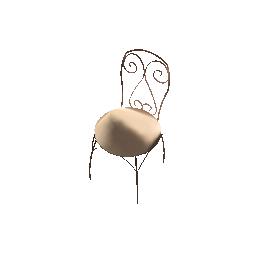} &
	\includegraphics[width=0.16\linewidth,trim={.75cm 1.75cm .75cm .75cm},clip]{gfx/singleview/0217_baseline_single_glob_resnet_geo_10lay_new/rot_light/shape08079/00000004_white.jpg} &
	\includegraphics[width=0.16\linewidth,trim={.75cm 1.75cm .75cm .75cm},clip]{gfx/singleview/0217_glob_resnet_geo/rot_light/shape08079/00000004_white.jpg} &
	\includegraphics[width=0.16\linewidth,trim={.75cm 1.75cm .75cm .75cm},clip]{gfx/singleview/0217_glob_resnet_geo/rot_light/shape08079/0001_0004_white.jpg} \\
\end{tabular}

    \caption{\textbf{Single View Appearance Prediction.} Qualitative comparison of our models (1-step, 2-step) trained on two (2 views) and all training views per objects. 
    While the models trained on two views predict reasonable texture, they show less accurate shadows and specular reflections.}
	\label{fig:2TV} 
\end{figure}

A qualitative comparison of our models trained on only two views with one light configuration and ten views with four light configurations for each view is depicted in \figref{fig:2TV}. 
The models trained on two training views show less accurate shadows and reflections compared to the models trained on all views.
However, even when using only two views per object during training our model is able to capture the coarse overall appearance of the object.

\end{document}